%% file: main.tex
\documentclass{article} %
\input{preamble}

\title{MosaiChunk: Compositing Spatio-Temporal\\ Memory for Autoregressive Video Generation}

\author{%
\begin{minipage}[t]{\textwidth}
\centering\normalfont
Yiwen Zhang$^{1,*}$ \quad Haocheng Xi$^{2}$ \quad
Michael Tian-Yue Liu$^{2,4}$ \quad Alexei A. Efros$^{2}$\\
Hadar Averbuch-Elor$^{1}$ \quad Qianqian Wang$^{3,4,\dagger}$ \quad
Haiwen Feng$^{2,4,\dagger}$\\[0.5em]
{\small
$^{1}$Cornell University \quad
$^{2}$University of California, Berkeley\\
$^{3}$Harvard University \quad
$^{4}$Impossible, Inc.\\[0.25em]
$^{*}$Part of the work done during an internship at Impossible Inc.\\
$^{\dagger}$Equal supervision.}\\[0.6em]
{\hypersetup{hidelinks}\href{https://mosaichunk.github.io/}{%
\textcolor[HTML]{327AC0}{\texttt{https://mosaichunk.github.io/}}}}
\end{minipage}
}

\iclrfinalcopy
\begin{document}

{\setlength{\tabcolsep}{0pt}\maketitle}
\ificlrfinal\lhead{Preprint}\fi

\input{sec/0_abstract}

\input{sec/1_intro_beta}
\input{sec/2_related}
\input{sec/3_method}
\input{sec/4_benchmark}
\input{sec/5_results}
\input{sec/6_conclusion}

\input{sec/7_statements}

\bibliography{main}
\bibliographystyle{iclr2027_conference}

\input{sec/A_appendix}

\end{document}

%% file: preamble.tex
\usepackage[T1]{fontenc}
\usepackage{iclr2027_conference,times}

\input{math_commands.tex}

\usepackage{hyperref}
\usepackage{url}

\usepackage{graphicx}
\usepackage{wrapfig}
\usepackage{needspace}
\usepackage{booktabs}
\usepackage{amsmath,amssymb}
\usepackage{pifont}
\usepackage{subcaption}
\usepackage{placeins}
\usepackage{multirow}
\usepackage{xcolor}
\usepackage{xspace}
\usepackage{textcomp}
\usepackage{upquote}

\graphicspath{{figures/}}

\newcommand{\ignorethis}[1]{{}}

\newcommand{\parnobf}[1]{%
  \par\smallskip%
  \noindent\textbf{#1}\hspace{0.4em}\ignorespaces}

\newcommand{\method}{\textit{MosaiChunk}\xspace}
\newcommand{\methodpose}{\textit{MosaiChunk-Pose}\xspace}
\newcommand{\bench}{\textit{RememBench}\xspace}

%% file: math_commands.tex
\usepackage{amsmath,amsfonts,bm}

\def\eqref#1{equation~\ref{#1}}

\def\1{\bm{1}}

\DeclareMathAlphabet{\mathsfit}{\encodingdefault}{\sfdefault}{m}{sl}
\SetMathAlphabet{\mathsfit}{bold}{\encodingdefault}{\sfdefault}{bx}{n}

%% file: sec/0_abstract.tex
\begingroup
\newsavebox{\teaserbox}
\sbox{\teaserbox}{\begin{minipage}{\textwidth}
\centering
\includegraphics[width=\textwidth,trim=0bp 8bp 0bp 18bp,clip]{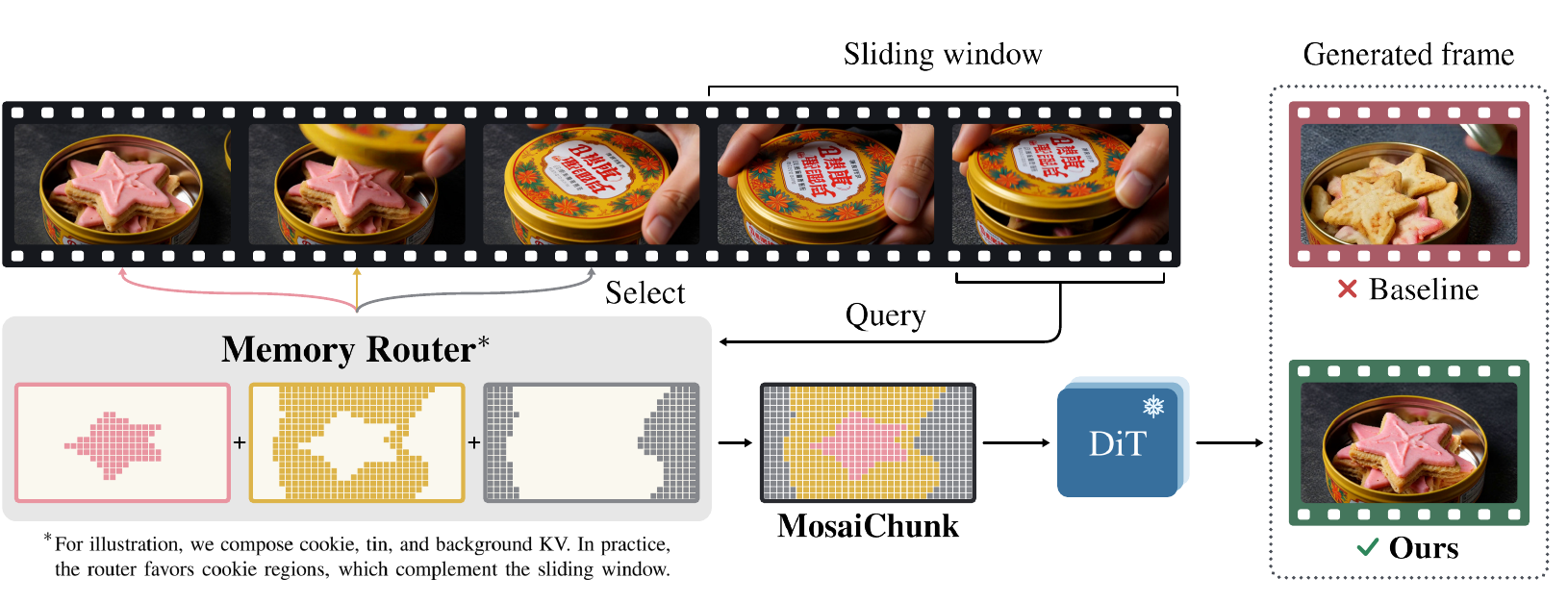}
\captionof{figure}{
In text-to-video generation, prompts like {\em ``Close the tin, then reopen it.''} illustrate a well-known problem: once a frame (e.g. tin with cookie) is dropped from the context sliding window, its contents are forgotten and must be hallucinated (top). 
We introduce \method (bottom), a spatio-temporal memory mechanism that conditions a
video generation model on composed historical KV sections (pink for the cookie,
yellow for the tin, and gray for the background). Our mechanism preserves fine-grained visual
details over long-horizon rollouts, such as the star cookie's appearance when
the tin closes and reopens.
}
\label{fig:teaser}
\end{minipage}}
\par\noindent\usebox{\teaserbox}\par
\endgroup

\begin{abstract}

Long-horizon autoregressive video generation is limited by a finite context window. When an object or scene falls out of context, its fine-grained visual details may be lost and difficult to recover upon reappearance. To retain access to such visual details, we introduce
\method, a spatio-temporal memory mechanism that composes a mosaic of selected historical key--value (KV) entries across space and time. Our approach is motivated by the observation that a frozen video generator can directly consume such non-contiguous historical KV and recover the corresponding visual content. We therefore keep the generator fixed and learn only a lightweight router that determines which historical sections to include in the mosaic under a fixed active-memory budget.
We further introduce \bench, a benchmark of long-horizon revisits with
prompt-driven text-to-video (T2V) and camera-driven image-to-video (I2V)
splits. Our experiments show that \method{} consistently improves revisit
consistency over both sliding-window inference and whole-chunk retrieval
under matched memory budgets, across both T2V and I2V settings.
All code and data will be released.
\end{abstract}

%% file: sec/1_intro_beta.tex
\section{Introduction}
\label{sec:intro}

\begin{quote}
\small
\textbf{``Nothing comes from nothing.''}\\
\hfill---\emph{The Sound of Music} (1965)
\end{quote}

The visual world is full of recurrence: objects return, surfaces reappear from new viewpoints, and details established at one moment become relevant again much later. For decades, vision and graphics methods have exploited this structure by constructing new imagery from visual elements that have already been observed---reusing patches, frames, and regions rather than synthesizing every detail from scratch~\citep{schodl2000videotextures,efros2001quilting,jojic2003epitomic,isola2013scene}. In this work, we ask: can the same principle of visual reuse support persistent memory in autoregressive video generation?

Consider the example illustrated in Figure \ref{fig:teaser}, where a video model is prompted to show a biscuit tin, close it, and reopen it a few seconds later. When the tin reopens, the model should be able to reuse the biscuit's previously generated appearance, since it has already established what is inside. Yet long-horizon autoregressive video generation typically relies on sliding-window inference, which keeps the active cache bounded by retaining only the most recent entries and evicting older ones. Once the biscuit leaves this window, the generator loses direct access to that specific appearance, and may synthesize a different biscuit when the tin reopens. %

The simplest solution would be to keep the entire generated history active, but this quickly becomes impractical: as the rollout grows, so do both the KV cache and the cost of attending to it. The challenge is therefore not to remember everything, but to recover a small set of historical KV needed for the current continuation. Recent approaches address this problem by retrieving the most relevant historical chunks, typically according to similarity with the current context~\citep{cai2025mixturecontextslongvideo,yi2026worldkv}. However, chunks bundle the keys and values of multiple frames, while the reusable visual content may occupy only a small subset of entries within these chunks.

Accordingly, we introduce \method, which constructs a new kind of memory ``chunk'' by assembling a mosaic of selected historical keys and values that come from different spatial regions and different moments in the generated history. For example, historical KV sections corresponding to the biscuit are composited with more recent sections corresponding to the tin and background in Figure~\ref{fig:teaser}. Our memory mechanism is enabled by the observation that a frozen video generator can already consume such non-contiguous historical KV and use it to recover the corresponding visual content. This allows us to keep the generator fixed and train only a lightweight router to retrieve and compose the historical sections most relevant to each continuation, thereby preserving previously established visual details under a fixed active-memory budget.

Because temporally neighboring sections can have similar keys, direct similarity-based retrieval can favor recent content and miss older content needed when previously seen objects or scenes reappear. 
To reduce this recency bias, we train the router's descriptor encoder through self-distillation, 
where both teacher and student use the same frozen backbone and noisy latent, differing only in their far memory. The teacher receives oracle memory containing whole departure chunks with the content to be redrawn; the student receives a smaller memory assembled by the router. Matching their denoising velocity predictions trains the descriptors that determine section selection and weighting. Because the targets come from the backbone's own predictions on rollout latents, this oracle-memory distillation requires no rendered or recorded ground-truth target videos.

To evaluate \method, we introduce \bench, a benchmark of long-horizon \emph{revisits}, in which previously seen content comes back into view. Such revisits provide a concrete test of a video model's memory: the returning content should preserve the details established earlier in the rollout. Autoregressive video models such as RAVEN-adapted MiniMax-H3 (H3-AR)~\citep{minimax2026h3,lu2026ravenrealtimeautoregressivevideo} and LingBot-World-Infinity~\citep{gao2026infiniteworldsversatileinteractions} support long interactive rollouts through prompt updates and camera controls. We leverage these interfaces to construct complementary prompt-driven T2V and camera-driven I2V splits. Compared with an existing whole-chunk retrieval and conditioning baseline, \method improves median revisit CLIP similarity from $0.841$ to $0.936$ on T2V and from $0.807$ to $0.863$ on I2V after the camera turns $180^\circ$ away and then back.

In summary, our contributions include: (i) a compositional visual memory mechanism that retrieves historical KV sections  under a fixed
active budget; (ii) a self-distillation strategy for learning KV section descriptions without
rendered ground-truth target videos; and (iii) \bench, a benchmark of
long-horizon revisits with prompt-driven T2V and camera-driven I2V splits.

%% file: sec/2_related.tex
\section{Related Work}
\label{sec:related}

\parnobf{Autoregressive video generation.}
Recent work has developed autoregressive video diffusion models for streaming and long-horizon generation. Diffusion Forcing~\citep{chen2024diffusionforcing} combines causal prediction with independently noised tokens, while CausVid~\citep{yin2025causvid} distills a bidirectional teacher into a few-step causal generator. Subsequent methods improve rollout training: Self Forcing~\citep{huang2025selfforcing} uses self-generated histories, MV-Forcing~\citep{fiebelman2026mv} extends self-forcing to joint temporal and view-wise autoregression, RAVEN~\citep{lu2026ravenrealtimeautoregressivevideo} allows later-chunk losses to supervise historical representations, and LongLive~\citep{yang2025longlive} trains on longer rollouts with prompt transitions. 
Sliding-window inference nevertheless evicts historical KV needed for revisits. We address this through learned retrieval under a fixed active cache budget.

\parnobf{Long-term visual memory.}
Prior methods extend visual memory through frame retrieval or geometric organization of past observations~\citep{yu2025contextasmemory,xiao2025worldmem,wu2025spatialmem}, or through compressed and learned memory representations~\citep{zhang2025framepack,relic2025,wu2025packforce}. Within native KV representations, cache-management methods reduce storage and attention costs through quantization, token selection, merging, and compaction~\citep{xi2026quantvideogen,chen2026pafukv,luo2026futureforcing,ji2026forcingkv,li2026packcache,yi2025deepforcing}.

Retrieval-based methods select historical context using content similarity, camera/action correspondence, or learned selection mechanisms, retrieving frames, chunks, or blocks to condition subsequent generation~\citep{cai2025mixturecontextslongvideo,yi2026worldkv,wu2026addressablememoryvideoworld,zhao2026omnimem}. Training these memory mechanisms can involve jointly optimizing memory selection and the generator, or fine-tuning the generator on revisit sequences~\citep{zhao2026omnimem,xue2026ringforcingpreciselongterm,chen2026pafukv}. In contrast, \method{} retrieves fine-grained, content-based sections of original KV under a fixed active cache budget. We keep the backbone frozen and train only the selector using denoising predictions under oracle context, without rendered ground-truth target videos.

\parnobf{Representing visual content with reusable components.}
Earlier work synthesizes images and videos by reusing visual fragments, from texture patches and video frames~\citep{efros2001quilting,schodl2000videotextures} to regions retrieved for completion and scene composition~\citep{wexler2004spacetime,hays2007scene,isola2013scene}. Related approaches capture recurring appearance through epitomic representations~\citep{jojic2003epitomic}, discover objects from local-feature statistics~\citep{sivic2005discovering}, or measure similarity through composition~\citep{boiman2006similarity}.
We adopt this perspective of decomposing visual content into reusable components, organizing historical KV into content-based sections that can be retrieved and composed across time.

%% file: sec/3_method.tex
\begin{figure}[t]
\centering
\includegraphics[width=\textwidth,trim=0bp 8bp 0bp 14bp,clip]{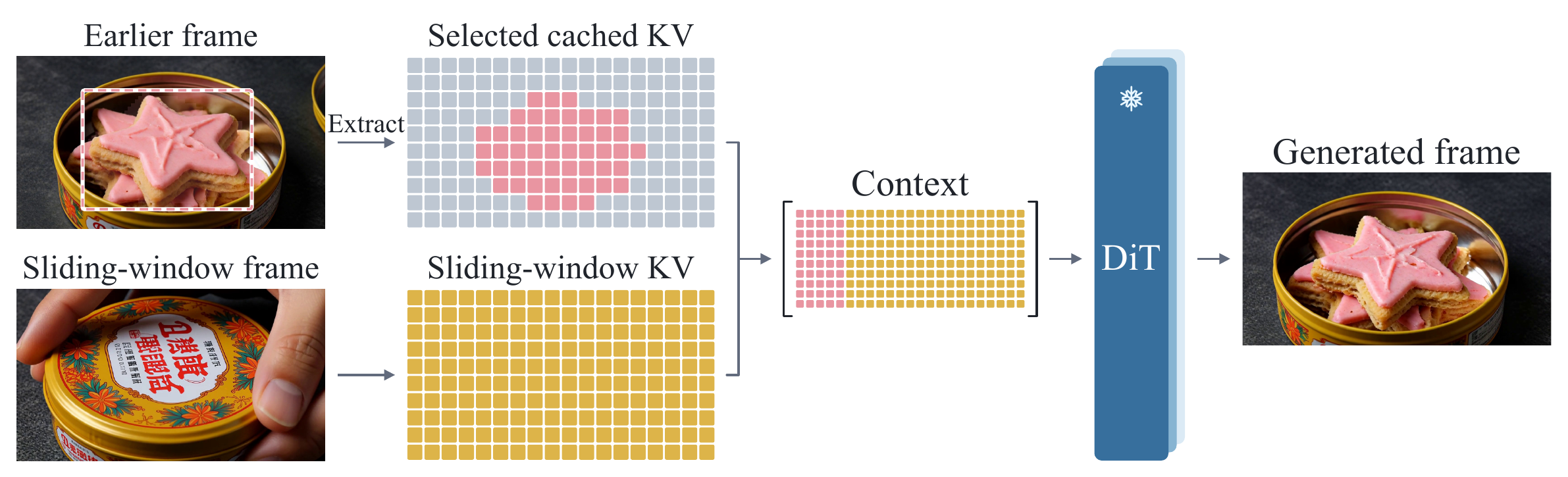}
\caption{\textbf{Can a frozen backbone reuse a non-contiguous subset of historical KV?}
To test whether a frozen video generator can consume a non-contiguous subset of KV entries, we select the KV at positions corresponding to the marked cookie
region. Only the colored entries are supplied alongside the sliding window.
The frozen video generator preserves the cookie's appearance when the tin opens again.}
\label{fig:feed}
\end{figure}

\section{Method}
\label{sec:method}

We first examine how a frozen video backbone can use non-contiguous
historical KV entries (Section~\ref{sec:method:probe}). This observation
motivates \method's memory architecture and the training strategy for its
descriptor encoder (Section~\ref{sec:method:bank}).

\subsection{Reusing Non-Contiguous Historical KV for Visual Memory}
\label{sec:method:probe}

Autoregressive video models generate one chunk at a time, attending to
cached keys and values (KV) from earlier chunks. Under sliding-window
inference, the backbone is conditioned on the most recent chunks, as well as a fixed attention sink used to stabilize generation; we omit this additional sink from the explanations below for simplicity.

To condition generation on more distant memory, existing retrieval methods select past chunks as additional context \citep{cai2025mixturecontextslongvideo,yi2026worldkv}. But are whole chunks necessary? We test whether a video generator can be conditioned on non-contiguous chunks, providing targeted KV sections as additional context. We select only the KV entries that represent the relevant content as far memory. We ask if these entries can steer the generation and bring that content back.

Figure~\ref{fig:feed} illustrates this intervention. We retain a
copy of an earlier chunk's KV before eviction and manually mark the cookie
region in one frame. We map this region to latent positions and select
the corresponding cached keys and values. We supply them alongside
the sliding window when the tin opens again. As illustrated in the figure, the generated frame preserves the cookie's pink, star-shaped appearance, demonstrating that a frozen autoregressive video generator can directly reuse a non-contiguous subset of its own previously cached KV, even after those entries have left the active context. This observation motivates our approach, detailed below, which reuses non-contiguous KV chunks for providing video generation with rich, yet minimal, historical context.

\subsection{Composing Historical KV with \method}
\label{sec:method:bank}
\label{sec:method:distill}

Section~\ref{sec:method:probe} shows that selecting non-contiguous historical KV entries
for the relevant content can produce a consistent rollout. Our goal is to automatically identify and compose the right historical KV entries under a fixed active-cache budget. In what follows, we first describe the architecture of our memory router. We then introduce our self-distillation scheme for learning a section descriptor space in which past sections are matched and retrieved.

\begin{figure}[t]
\centering
\includegraphics[width=\textwidth,trim=0bp 4bp 0bp 7bp,clip]{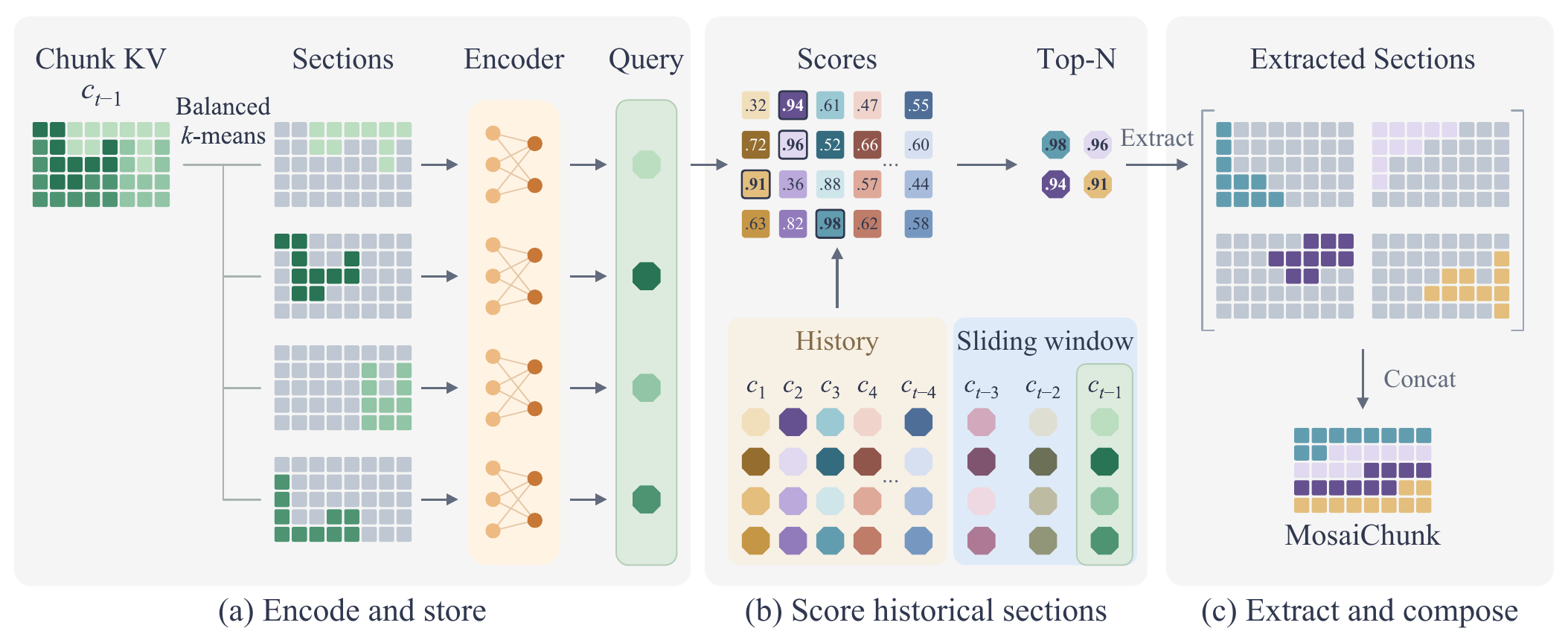}
\caption{\textbf{Memory architecture.}
(a) Each chunk is partitioned into sections and each section is encoded into a descriptor.
(b) Sections in the latest chunk query history outside the sliding window.
(c) The N sections with the highest scores are extracted and composed into a \method.}
\label{fig:bank}
\end{figure}

\parnobf{Memory architecture.}
Our \emph{memory router} follows a three-stage design: it encodes and stores
generated chunks as fine-grained sections, scores historical sections against
the latest query, and composes a \method from the selected sections; an overview is provided in Figure~\ref{fig:bank}.

\emph{Stage 1: Encode and store.}
Since we want to select groups of historical KV entries with some specific semantic meaning, we partition each chunk's unrotated KV into equal-size groups using balanced
$k$-means and call these groups \emph{sections}. We observe that the KV entries for each section exhibit strong temporal redundancy: sections from neighboring chunks are always similar.
Therefore, we train a lightweight \emph{descriptor encoder} that maps the pooled keys of each section $s$ to a
descriptor $d(s)$. The descriptor space learns an anti-recency bias so that similarity over descriptors reflects semantic relatedness rather than temporal proximity. The \emph{section bank} stores these sections of verbatim
KV entries and their descriptors in CPU memory.

\emph{Stage 2: Score historical sections.}
For retrieval when generating chunk $c_t$, the sections of the
latest generated chunk $c_{t-1}$ form the query set $\mathcal{Q}$, and the historical sections outside
the sliding window form the candidate set $\mathcal{H}$. The goal is to retrieve
sections from $\mathcal{H}$ given all sections in $\mathcal{Q}$.
We compute similarity between the descriptors of $\mathcal{Q}$ and $\mathcal{H}$.
The router scores each candidate $h\in\mathcal{H}$ as:
\begin{equation}
\operatorname{score}(h)
=\max_{q\in\mathcal{Q}}\;\cos\bigl(d(q),\,d(h)\bigr),
\qquad h\in\mathcal{H}.
\label{eq:score}
\end{equation}
The score measures a candidate's highest descriptor similarity to any query
section. High-scoring sections therefore retrieve historical content relevant
to the latest chunk.

\emph{Stage 3: Extract and compose.}
The router ranks all candidate sections together and selects the global
top-$N$. Since sections have equal size, the far-memory budget determines
$N$. A softmax over all candidate scores gives a weight for each section.
We normalize the selected weights to unit mean and use them to scale the
selected sections' values. We then concatenate their KV into a
\method, which serves as \emph{far memory} for chunk $t$.
The frozen DiT reads the \method alongside the sliding window. We sweep multiple values of $N$ in our evaluations, where a larger $N$ corresponds to a larger \method budget and broader coverage of the history.

\begin{figure}[t]
\centering
\includegraphics[width=\textwidth,trim=0bp 13bp 0bp 34bp,clip]{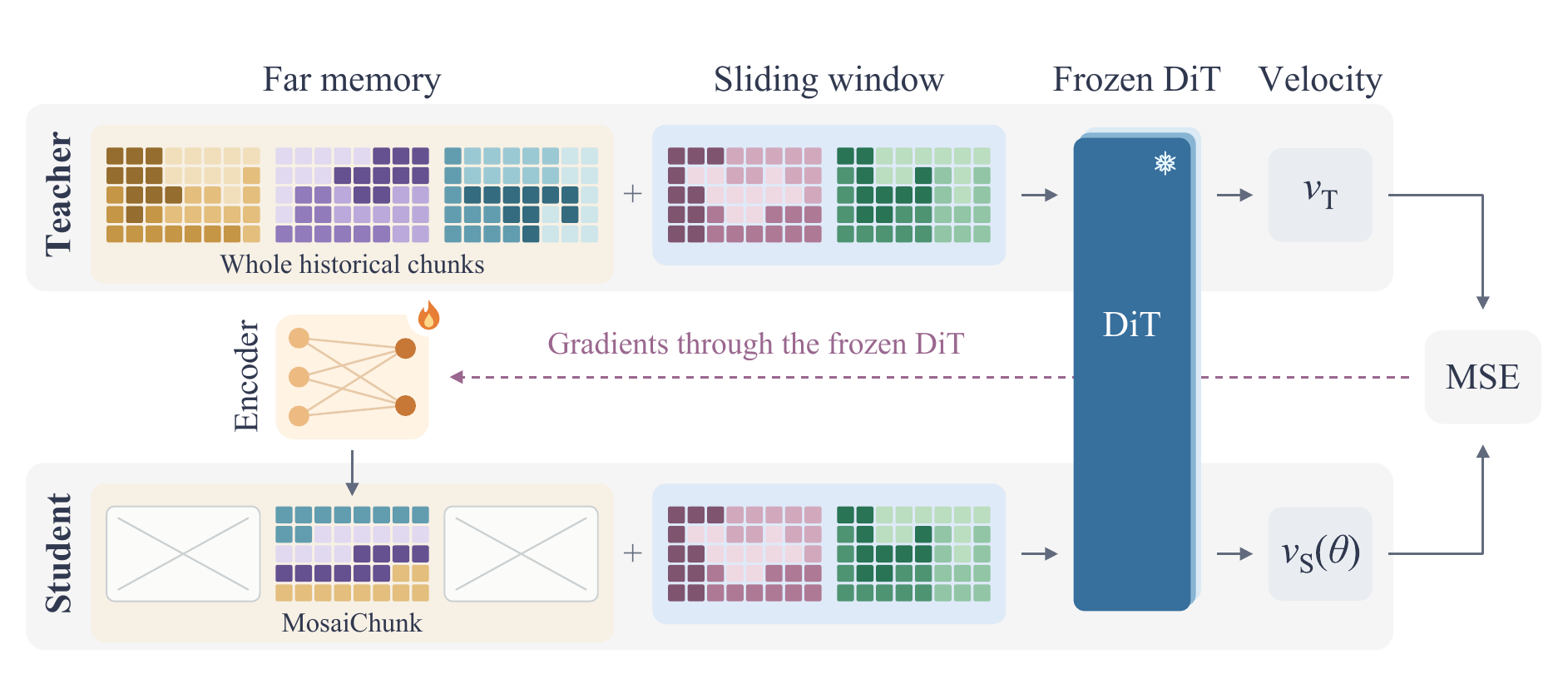}
\caption{\textbf{Training pipeline.}
The far memory consists of whole historical chunks for the teacher and a
smaller \method composed by our memory router for the student. The far memory is joined with the same sliding window and sent into the frozen DiT. Matching their velocity predictions
updates the descriptor space of the encoder.}
\label{fig:distill}
\end{figure}

\parnobf{Training the descriptor encoder.} We train the descriptor encoder through self-distillation, teaching a student conditioned on a composed \method to match the denoising velocity of the same frozen generator acting as a teacher with richer far memory that entails the historical content to be revisited.

Both the teacher and the student run one pass through the same frozen DiT with
the same noisy latent and sliding window. Only the far memory differs, as shown in Figure~\ref{fig:distill}.
The \emph{teacher} receives whole historical chunks that fully cover the
historical content to be redrawn. We identify these chunks from the input prompt schedule
for T2V models or matching input camera poses for I2V models.
The \emph{student} receives the router's \method under a smaller memory
budget. It must therefore select useful sections rather than copy all
of the teacher's context. Let $\theta$ denote the parameters of the descriptor
encoder, $v_T$ the teacher's denoising velocity prediction, and
$v_S(\theta)$ the student's prediction. We minimize their
mean squared error over training chunks and noise levels:
\begin{equation}
\mathcal{L}_{\mathrm{distill}}(\theta)
=\mathbb{E}\left[\left\lVert v_S(\theta)-v_T\right\rVert_2^2\right].
\label{eq:distill}
\end{equation}
The teacher prediction is a fixed target. Gradients pass through the
student's value weights to the descriptor encoder, not through the
discrete top-$N$ indices. The stored KV and backbone parameters remain
unchanged. Since the backbone itself supplies the target, training
requires no rendered ground-truth target videos.

Since teacher and student differ only in far memory, the self-distillation signal directly supervises memory selection and weighting. The teacher's whole chunks come from a much earlier visit to the same content, so matching its predictions encourages the student to recover useful details from older history. This encourages an anti-recency bias to emerge in the learned descriptor space: sections are favored for their relevant content rather than their recency alone.

%% file: sec/4_benchmark.tex
\section{Benchmarks}
\label{sec:benchmark}

Existing benchmarks such as WBench~\citep{ying2026wbench},
WorldMark~\citep{xu2026worldmark}, and PersistBench~\citep{he2026persistbench}
evaluate video quality and consistency,
but do not systematically test \emph{long-horizon revisits} in which frames establishing
the target content's appearance are evicted from the model's sliding window.
We therefore introduce \bench, a benchmark with two splits for evaluating
consistency across such revisits in autoregressive video models.
Revisits are driven by
prompts in the T2V split and by camera motion in the I2V split.
Additional benchmark construction details and first-frame visualizations are
provided in Supplementary Sections~\ref{app:construction} and~\ref{app:first_frames},
respectively.

\parnobf{The T2V split.}
The split contains 100 samples with scenarios disjoint from router training.
Each model input extends Ring Forcing's three-stage
appear--disappear--reappear design~\citep{xue2026ringforcingpreciselongterm}
to four prompt segments, with a separate segment keeping the object out of sight.

\parnobf{The I2V split.}
The split contains 150 scenes: 50 indoor and 100 outdoor. Each model input
includes an initial frame from DL3DV~\citep{ling2023dl3dv10klargescalescenedataset},
a prompt describing the scene, and a camera trajectory. We sample more diverse camera trajectories not seen during training: all 150 scenes have
in-place rotation trajectories with $90^\circ$, $180^\circ$, and $360^\circ$
settings. The 100 outdoor scenes additionally have trajectories combining
these rotations with translation.

%% file: sec/5_results.tex
\section{Results}
\label{sec:results}
\setcounter{topnumber}{3}
\setcounter{totalnumber}{4}

We evaluate whether \method provides useful fine-grained historical
information beyond a sliding window and compare it with existing
retrieval mechanisms that condition on whole chunks, testing whether
section-level memory better preserves visual details. We conduct both
comparisons on the T2V and I2V splits of \bench, using the frozen H3-AR
backbone for T2V and LingBot-World-Infinity for I2V.
Video comparisons are available in the \href{https://mosaichunk.github.io/viewer/video.html}{Video~Viewer}.

\input{tables/t2v}

\subsection{Evaluation Protocol}
\label{sec:results:protocol}

\parnobf{Baseline.} We compare \method with the sliding-window baseline (Base) and Mixture of
Contexts (MoC)~\citep{cai2025mixturecontextslongvideo}, a whole-chunk
retrieval mechanism. We adapt MoC on each frozen backbone. Across baselines, T2V runs share prompts and noise seeds;
I2V runs share conditioning frames and camera trajectories.

\parnobf{Notation.} Let $\mathcal{F}_c$ denote the far memory used to generate chunk $c$, and
$\lVert\mathcal{F}_c\rVert$ its budget in video-chunk equivalents. For example, $\lVert\mathcal{F}_c\rVert = 1$ means that the size of the far memory is equivalent to one video chunk. We evaluate
budgets of one and two chunks.

\parnobf{Metric.} CLIP similarity and LPIPS assess consistency between departure and revisit
frames. T2V pairs are manually marked; I2V pairs use the conditioning frame
and a revisit selected from Pi3X-reconstructed camera
poses~\citep{wang2026pi3permutationequivariantvisualgeometry}.
TempSSIM (consecutive-frame SSIM~\citep{wang2004ssim}) and Local Scene Drift
(Drift)~\citep{wu2026addressablememoryvideoworld} measure rollout quality.
We report medians over scenes. More details are provided in Supplementary Material
Section~\ref{app:evaluation}.

\input{tables/i2v}

\begin{figure}[tb]
\centering
\includegraphics[width=0.95\textwidth]{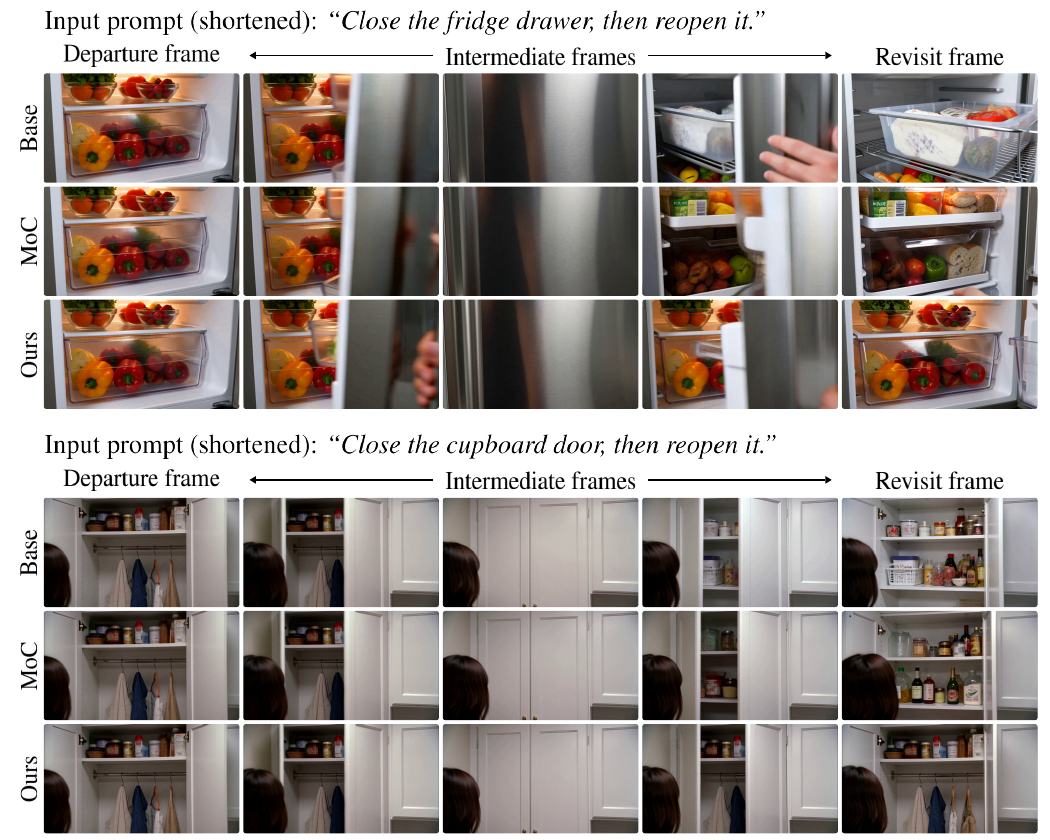}
\caption{\textbf{T2V revisits at a two-chunk far-memory budget.}
Rows (Base, MoC, and Ours) share the prompt and noise within each scene.
The first and last columns show the departure and revisit frames selected
for evaluation.
Only \method preserves the vegetables and their arrangement (top) and the
cupboard's contents (bottom).}
\label{fig:qual}
\end{figure}

\subsection{Text-to-Video Results}
\label{sec:results:t2v}

Table~\ref{tab:t2v} shows that \method improves median CLIP over Base by
$0.144$ and $0.185$ at budgets of one and two chunks, respectively, and over
MoC by $0.067$ and $0.095$. LPIPS shows the same ordering. These results
support allocating the retrieval budget to content-relevant sections rather
than whole chunks. At each budget, TempSSIM and Drift differ by less than
$0.01$ across methods, indicating similar rollout quality under these
metrics.
Figure~\ref{fig:qual} illustrates these differences at the two-chunk budget.
Only \method preserves the vegetables and their arrangement after the
refrigerator reopens, and retains the cupboard's contents after its doors reopen.

\input{tables/public_benchmarks}

\begin{figure}[tb]
\centering
\includegraphics[width=0.95\textwidth]{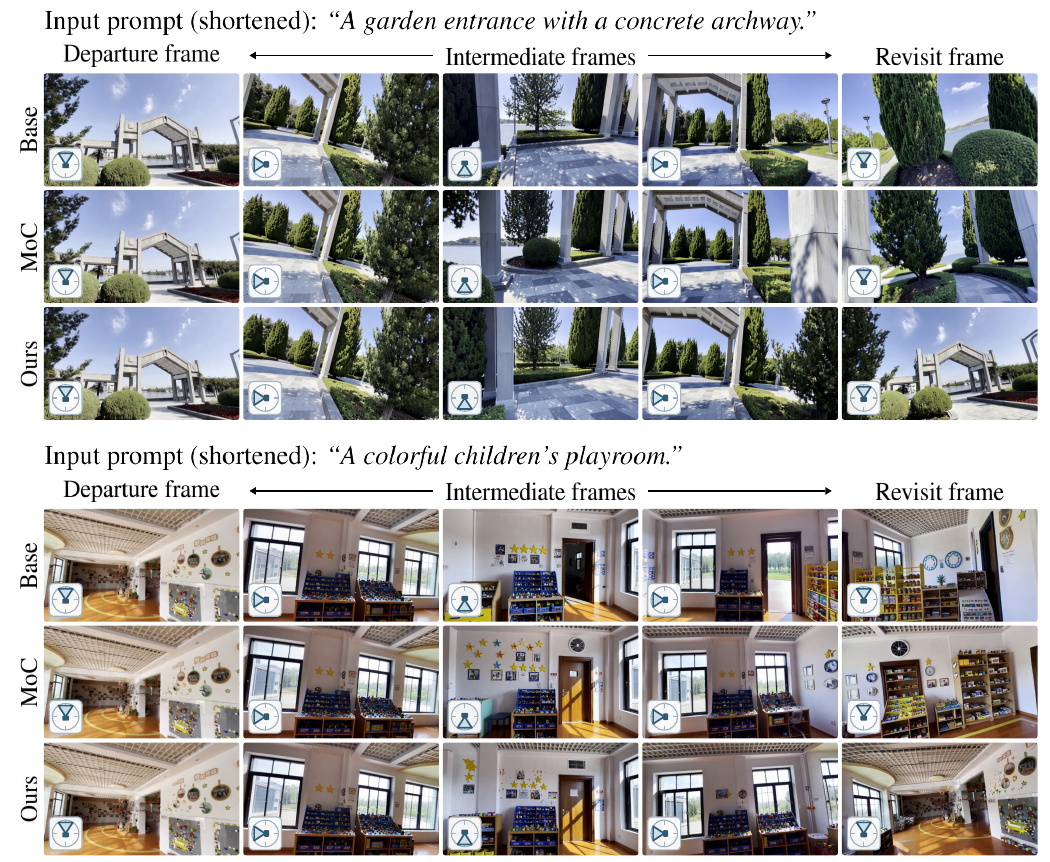}
\caption{\textbf{I2V revisits at two-chunk (top) and one-chunk (bottom) far-memory budgets.}
Rows (Base, MoC, and Ours) share the conditioning frame and $180^\circ$
camera trajectory within each scene. The first
column shows the departure frame; the last shows each method's revisit
frame, selected from its Pi3X reconstruction. Intermediate columns sample
the camera turn; camera insets schematically show the commanded rotation.
\method better recovers the garden entrance's archway (top)
and the playroom's layout and wall decorations (bottom).}
\label{fig:qual_i2v}
\end{figure}

\subsection{Image-to-Video Results}
\label{sec:results:i2v}

Table~\ref{tab:i2v} reports $180^\circ$ turns with and without translation.
At a one-chunk budget, \method improves median CLIP over Base by $0.072$
under rotation and $0.060$ with translation, and over MoC by $0.033$ and
$0.036$. At two chunks, the corresponding gains are $0.070$ and $0.049$
over Base, and $0.056$ and $0.032$ over MoC. LPIPS again improves in all
four settings. At each budget and trajectory, TempSSIM and Drift differ by
less than $0.01$ across methods.
Results for $90^\circ$ and $360^\circ$ turns are provided in Supplementary
Material Section~\ref{app:quantitative}.
Figure~\ref{fig:qual_i2v} shows two $180^\circ$ examples in which \method
more faithfully recovers the garden entrance's archway and the playroom's layout and
wall decorations than either baseline.

In addition, we evaluate I2V generation on two public benchmarks,
WBench~\citep{ying2026wbench} and WorldMark~\citep{xu2026worldmark}
(Table~\ref{tab:public_benchmarks}). At both budgets, \method improves every
consistency score in Table~\ref{tab:public_benchmarks} over the sliding-window
baseline while maintaining video quality. Full results and metric descriptions
are provided in Supplementary Material Section~\ref{app:quantitative}.

\subsection{Ablation Studies}
\label{sec:results:ablation}

We ablate two key design choices in our memory architecture:
(1) learning a descriptor space for retrieval and (2) selecting sections
globally under a shared memory budget. All variants use the same section partition and frozen
backbone, evaluated on 150 I2V rotation scenes at one- and two-chunk
far-memory budgets (Table~\ref{tab:architecture}).

\input{tables/architecture}

\emph{Learning the descriptor space.}
As discussed in Section~\ref{sec:method:bank}, temporally neighboring
sections have similar keys. Directly matching pooled keys can therefore favor
recent sections over older content needed at a revisit. We train the
descriptor encoder to counter this recency bias.
We replace learned descriptors with mean-pooled keys and score their cosine
similarity, keeping global top-$N$ selection and using uniform value weights.
The learned router achieves higher CLIP and lower LPIPS at both budgets.

\emph{Selecting sections under a shared budget.}
We retrain the router with MoC-style per-query top-$k$
selection~\citep{cai2025mixturecontextslongvideo} under the same memory budget.
Intuitively, content that never leaves view may remain in the sliding
window, so retrieving $k$ sections for every query can spend budget on
information the backbone already has; global selection instead devotes the
budget to relevant content missing from local context.
Global top-$N$ again improves both metrics at both budgets, making the full
design the best-performing variant in Table~\ref{tab:architecture}.

Since camera poses are available in I2V, we also test a pose-guided variant
of our router, following pose-based retrieval methods~\citep{yi2026worldkv}.
We further evaluate fractional far-memory budgets of $0.5$ and $1.5$ chunks
and report the budget curve. Both studies are provided in Supplementary
Material Section~\ref{app:ablation}.
\FloatBarrier

%% file: tables/t2v.tex
\begin{table}[tb]
\centering
\small
\setlength{\tabcolsep}{3pt}
\renewcommand{\arraystretch}{1.08}
\caption{\textbf{T2V results on \bench.}
Median CLIP and LPIPS on 100 marked departure--revisit pairs, with rollout-quality
metrics alongside. Budgets are in chunk equivalents; Base uses the same total
active KV cache budget. Best CLIP and LPIPS at each budget are bold.}
\label{tab:t2v}
\begin{tabular*}{0.9\textwidth}{@{\extracolsep{\fill}}lcccc@{\hspace{12pt}}cccc@{}}
\toprule
& \multicolumn{4}{c}{$\lVert\mathcal{F}_c\rVert=1$}
& \multicolumn{4}{c}{$\lVert\mathcal{F}_c\rVert=2$} \\
\cmidrule(lr){2-5}\cmidrule(lr){6-9}
Method & CLIP $\uparrow$ & LPIPS $\downarrow$ & TempSSIM $\uparrow$ & Drift $\downarrow$
       & CLIP $\uparrow$ & LPIPS $\downarrow$ & TempSSIM $\uparrow$ & Drift $\downarrow$ \\
\midrule
Base & 0.755 & 0.659 & 0.923 & 0.051 & 0.751 & 0.652 & 0.926 & 0.050 \\
MoC & 0.832 & 0.594 & 0.928 & 0.052 & 0.841 & 0.569 & 0.930 & 0.053 \\
Ours & \textbf{0.899} & \textbf{0.567} & 0.932 & 0.052 & \textbf{0.936} & \textbf{0.500} & 0.932 & 0.055 \\
\bottomrule
\end{tabular*}
\end{table}

%% file: tables/i2v.tex
\begin{table}[tb]
\centering
\small
\setlength{\tabcolsep}{3pt}
\renewcommand{\arraystretch}{1.08}
\caption{\textbf{I2V results on \bench.}
Median revisit CLIP and LPIPS for $180^\circ$ turns: 150 scenes with rotation
and 100 with translation. Budget and bolding conventions follow Table~\ref{tab:t2v}.}
\label{tab:i2v}
\begin{tabular*}{0.9\textwidth}{@{\extracolsep{\fill}}lcccc@{\hspace{12pt}}cccc@{}}
\toprule
& \multicolumn{4}{c}{$\lVert\mathcal{F}_c\rVert=1$}
& \multicolumn{4}{c}{$\lVert\mathcal{F}_c\rVert=2$} \\
\cmidrule(lr){2-5}\cmidrule(lr){6-9}
Method & CLIP $\uparrow$ & LPIPS $\downarrow$ & TempSSIM $\uparrow$ & Drift $\downarrow$
       & CLIP $\uparrow$ & LPIPS $\downarrow$ & TempSSIM $\uparrow$ & Drift $\downarrow$ \\
\midrule
\multicolumn{9}{@{}l}{\textit{Rotation}} \\[1pt]
Base & 0.768 & 0.655 & 0.457 & 0.051 & 0.793 & 0.650 & 0.454 & 0.052 \\
MoC & 0.807 & 0.645 & 0.457 & 0.050 & 0.807 & 0.642 & 0.460 & 0.049 \\
Ours & \textbf{0.840} & \textbf{0.625} & 0.456 & 0.051 & \textbf{0.863} & \textbf{0.609} & 0.458 & 0.051 \\
\midrule
\multicolumn{9}{@{}l}{\textit{Rotation + translation}} \\[1pt]
Base & 0.788 & 0.625 & 0.527 & 0.043 & 0.813 & 0.611 & 0.521 & 0.044 \\
MoC & 0.812 & 0.617 & 0.521 & 0.045 & 0.830 & 0.616 & 0.521 & 0.046 \\
Ours & \textbf{0.848} & \textbf{0.606} & 0.519 & 0.044 & \textbf{0.862} & \textbf{0.584} & 0.513 & 0.046 \\
\bottomrule
\end{tabular*}
\end{table}

%% file: tables/public_benchmarks.tex
\begin{table}[tb]
\centering
\small
\setlength{\tabcolsep}{2pt}
\renewcommand{\arraystretch}{1.08}
\caption{\textbf{I2V results on public benchmarks.}
WBench uses the navigation split; WorldMark uses the first-person splits.
Budgets specify \method's far memory in chunk equivalents; Base matches
the total active KV cache budget. Higher is better
for all metrics; bold marks the better consistency score at each budget.}
\label{tab:public_benchmarks}
\begin{tabular*}{0.9\textwidth}{@{\extracolsep{\fill}}cl*{9}{c}@{}}
\toprule
& & \multicolumn{6}{c}{WBench} & \multicolumn{3}{c}{WorldMark} \\
\cmidrule(lr){3-8}\cmidrule(lr){9-11}
& & \multicolumn{4}{c}{Consistency} & \multicolumn{2}{c}{Video quality}
& Consistency & \multicolumn{2}{c}{Video quality} \\
\cmidrule(lr){3-6}\cmidrule(lr){7-8}\cmidrule(lr){9-9}\cmidrule(lr){10-11}
$\lVert\mathcal{F}_c\rVert$ & Method & Spatial & \shortstack{Gated\\Spatial}
& Geom. & Subject & Aesth. & Imaging & \shortstack{Revisit\\Memory} & Aesth. & Percept. \\
\midrule
\multirow{2}{*}{1} & Base & 78.28 & 75.47 & 85.98 & 88.13 & 61.75 & 67.31 & 80.14 & 61.43 & 85.49 \\
& Ours & \textbf{79.63} & \textbf{76.69} & \textbf{86.74} & \textbf{88.36} & 61.86 & 67.42 & \textbf{84.89} & 62.48 & 86.45 \\
\midrule
\multirow{2}{*}{2} & Base & 79.43 & 76.78 & 86.14 & 88.10 & 61.76 & 67.40 & 84.63 & 61.05 & 84.97 \\
& Ours & \textbf{80.73} & \textbf{78.18} & \textbf{87.08} & \textbf{88.38} & 61.92 & 67.45 & \textbf{84.87} & 62.89 & 86.64 \\
\bottomrule
\end{tabular*}
\end{table}

%% file: tables/architecture.tex
\begin{table}[htb]
\centering
\small
\setlength{\tabcolsep}{5pt}
\renewcommand{\arraystretch}{1.15}
\caption{\textbf{Memory architecture ablation.}
Median revisit CLIP and LPIPS on 150 I2V rotation scenes.
Disabling descriptor learning uses mean-pooled keys; disabling global
selection uses per-query top-$k$. The last row is the full \method router,
with results from Table~\ref{tab:i2v}.}
\label{tab:architecture}
\begin{tabular*}{0.9\textwidth}{@{\extracolsep{\fill}}cccc@{\hspace{10pt}}cc@{}}
\toprule
\multicolumn{2}{c}{Router design}
& \multicolumn{2}{c}{$\lVert\mathcal{F}_c\rVert=1$}
& \multicolumn{2}{c}{$\lVert\mathcal{F}_c\rVert=2$} \\
\cmidrule(lr){1-2}\cmidrule(lr){3-4}\cmidrule(lr){5-6}
Learned descriptors & Global top-$N$
& CLIP $\uparrow$ & LPIPS $\downarrow$
& CLIP $\uparrow$ & LPIPS $\downarrow$ \\
\midrule
\ding{56} & \ding{52} & 0.815 & 0.637 & 0.841 & 0.640 \\
\ding{52} & \ding{56} & 0.792 & 0.656 & 0.792 & 0.655 \\
\midrule
\ding{52} & \ding{52}
& \textbf{0.840} & \textbf{0.625} & \textbf{0.863} & \textbf{0.609} \\
\bottomrule
\end{tabular*}
\end{table}

%% file: sec/6_conclusion.tex
\section{Conclusion}
\label{sec:conclusion}
We presented \method, a spatio-temporal memory mechanism for preserving
visual content across long-horizon revisits, and \bench for evaluating
this capability. Our self-distillation scheme trains the router using
the same frozen backbone with richer historical context as the teacher.
While the router improves visual memory, generation remains constrained
by the teacher's capabilities. Because the backbone's weights remain fixed,
our method inherits its limitations in instruction following and its
learned generative distribution. Future video models could instead build memory directly into the generator, learning during pre-training or post-training to natively reuse visual content from earlier in the generated history, even after it has left the active context.

%% file: sec/7_statements.tex
\subsection*{AI use statement}
We used generative AI tools to assist with data construction, research-code
implementation and debugging, interpreting experimental results, and manuscript preparation. Specifically for data generation, a
Claude agent wrote T2V prompt scenarios, while Qwen2.5-VL-7B assisted with
I2V scene filtering and captioning, as described in
Supplementary Section~\ref{app:construction}. An author manually reviewed all
AI-assisted text, code, data, and output visualizations. The
authors remain responsible for the accuracy and integrity of the methods,
results, and conclusions reported in this work. 

\subsection*{Ethics statement}
\method trains only a lightweight memory router while keeping the pretrained
video backbone frozen. It selects and composes historical KV entries without
updating the generator's parameters. The method therefore inherits
limitations of its pretrained backbone and source data, including potential
biases and harmful behaviors. We do not identify additional ethical
concerns specific to the proposed memory router.

\subsection*{Reproducibility statement}
Section~\ref{sec:method} describes the memory architecture and the
self-distillation objective. Supplementary Sections~\ref{app:implementation}
and~\ref{app:training} provide implementation and training details,
including hyperparameters and hardware. Supplementary
Section~\ref{app:construction} documents the data sources, filtering
procedures, prompts, and camera trajectories used to construct \bench,
while Section~\ref{app:evaluation} specifies generation settings, baseline
adaptations, matched memory budgets, revisit-pair selection, and metrics.
We will publicly release our code, data, and trained router checkpoints
to facilitate reproduction and further research.

%% file: sec/A_appendix.tex
\clearpage
\appendix
\raggedbottom
\renewcommand{\thefigure}{\arabic{figure}}
\renewcommand{\thetable}{\arabic{table}}
\renewcommand{\theequation}{\arabic{equation}}
\makeatletter
\setlength{\@fptop}{0pt}
\makeatother

\begin{center}
{\Large\bfseries Supplementary Material}
\end{center}

We refer readers to the interactive visualizations in the
\href{https://mosaichunk.github.io/viewer/video.html}{Video~Viewer}, which show text-to-video (T2V) and image-to-video (I2V)
results. In this document,
we provide implementation details (Section~\ref{app:method}), describe
benchmark construction, scene visualizations, and evaluation protocols
(Section~\ref{app:benchmarks}), and present additional results
(Section~\ref{app:results}).

\begingroup
\hypersetup{linkcolor=black}
\newcommand{\suppcontentssection}[2]{%
  \par\noindent\begingroup\bfseries
  \makebox[1.6em][l]{\ref{#1}}\hyperref[#1]{#2}%
  \hfill\makebox[1.5em][r]{\pageref{#1}}\par\endgroup}
\newcommand{\suppcontentssubsection}[2]{%
  \par\noindent\hspace*{1.6em}%
  \makebox[2.4em][l]{\ref{#1}}\hyperref[#1]{#2}\nobreak\hspace{0.5em}%
  \leaders\hbox to 0.5em{\hss.\hss}\hfill\nobreak
  \makebox[1.5em][r]{\pageref{#1}}\par}
\noindent\makebox[\textwidth][c]{%
\begin{minipage}{0.8\textwidth}
\setlength{\parskip}{0pt}
\textbf{Contents}\par\smallskip
\suppcontentssection{app:method}{Implementation Details}
\suppcontentssubsection{app:implementation}{Memory Router}
\suppcontentssubsection{app:training}{Self-Distillation}
\smallskip
\suppcontentssection{app:benchmarks}{Benchmark Details}
\suppcontentssubsection{app:construction}{Dataset Construction}
\suppcontentssubsection{app:first_frames}{Visualization of First Frames}
\suppcontentssubsection{app:evaluation}{Detailed Evaluation Protocol}
\smallskip
\suppcontentssection{app:results}{Additional Results}
\suppcontentssubsection{app:qual}{Extra Qualitative Results}
\suppcontentssubsection{app:quantitative}{Extra Quantitative Results}
\suppcontentssubsection{app:ablation}{Extra Ablations}
\suppcontentssubsection{app:kv_reuse}{KV Transplantation}
\end{minipage}}\par
\endgroup

\input{sec/appendix/A_method}
\input{sec/appendix/B_benchmarks}
\input{sec/appendix/C_results}

%% file: sec/appendix/A_method.tex
\section{Implementation Details}
\label{app:method}

\method adds a trainable memory router to a frozen autoregressive video
generator. The router composes historical KV sections into a \method
under a fixed far-memory budget. The frozen generator reads it alongside
a sliding window of two recent chunks and a fixed \emph{attention sink}.
We use RAVEN-adapted MiniMax-H3
(H3-AR) for text-to-video (T2V) and LingBot-World-Infinity for image-to-video
(I2V)~\citep{minimax2026h3,lu2026ravenrealtimeautoregressivevideo,gao2026infiniteworldsversatileinteractions}.
We observed content leakage through a visual sink in H3-AR, but not in
LingBot-World-Infinity. We therefore retain text tokens as the attention
sink for H3-AR and the first video chunk's tokens for LingBot-World-Infinity.

\subsection{Memory Router}
\label{app:implementation}

\parnobf{Stage 1: Encode and store.}
We partition each chunk's unrotated KV into equal-size groups, called
\emph{sections}, using balanced $k$-means. Equal
section sizes let the number of selected sections determine the KV budget.
Clustering uses keys before rotary positional encoding (RoPE), at layer
$25$ of H3-AR's $50$ layers and layer $20$ of LingBot-World-Infinity's $40$ layers.
The same section partition applies to all layers; sections may span non-contiguous
positions and multiple frames. Table~\ref{tab:router_geometry} lists their
sizes and the layers used for descriptors.

\begin{table}[!htb]
\centering
\caption{\textbf{Memory router settings for Stage 1.} Token counts refer to
visual tokens only; layer indices are zero-based. Keys from the clustering
layer are used to partition each chunk into equal-size sections. Pooled
keys from the descriptor layers are used to compute each section's descriptor.}
\label{tab:router_geometry}
\small
\begin{tabular}{@{}lll@{}}
\toprule
 & T2V (H3-AR) & I2V (LingBot-World-Infinity) \\
\midrule
Transformer layers & $50$ & $40$ \\
Attention heads $\times$ head width & $56\times128$ & $40\times128$ \\
Key width across heads & $7168$ & $5120$ \\
Latent frames per chunk & $5$ & $4$ \\
Spatial patch grid & $24\times43$ & $30\times52$ \\
Visual tokens per chunk & $5160$ & $6240$ \\
Sections per chunk, $S$ & $40$ & $48$ \\
Tokens per section & $129$ & $130$ \\
Clustering layer & $25$ & $20$ \\
Descriptor layers & $12,25,38$ & $10,20,30$ \\
\bottomrule
\end{tabular}
\end{table}

To compute each section's descriptor, we pool its unrotated keys at layers
$12,25,38$ for H3-AR and $10,20,30$ for LingBot-World-Infinity.
Each token's keys across attention
heads form one vector. Elementwise mean, maximum, and minimum over the
section's tokens capture average responses and extremes that averaging
would lose~\citep{gao2025seerattention}. These three statistics at three
layers yield nine pooled key vectors. The \emph{descriptor encoder} normalizes
their scales, projects them to width $1024$, adds embeddings identifying
the source layer and statistic, and refines them with a shared residual
MLP (hidden width $2048$, GELU). Separate output projections give query
and key descriptor sets $d_Q(s)$ and $d_K(s)$, each with nine vectors,
so the same section can serve either role in comparison.

To keep full historical KV off the GPU, the \emph{section bank} stores
verbatim KV entries from all backbone layers in bfloat16 on the CPU, together
with each section's token indices. Stored history grows with the rollout;
the DiT's active KV budget remains fixed.

\parnobf{Stage 2: Score historical sections.}
When generating chunk $t$, sections of the latest generated chunk $t-1$
form the query set $\mathcal Q$. Historical sections outside the sliding
window and attention sink form the candidate set $\mathcal H$.

Before comparison, we center key descriptor vectors within $\mathcal H$
for each source layer and pooling statistic,
to remove components shared across sections. We normalize query and
centered key vectors to unit length for cosine similarity. For descriptor
sets $A$ and $B$, $\operatorname{sim}(A,B)$ averages the best cosine match
in $B$ for each of $A$'s nine vectors. Each candidate receives the score
\begin{equation}
s(h)=\max_{q\in\mathcal Q}\operatorname{sim}\bigl(d_Q(q),d_K(h)\bigr),
\quad h\in\mathcal H.
\label{eq:supp_score}
\end{equation}
This favors sections that match a current query while complementing the
content that is missing from the sliding window.

\parnobf{Stage 3: Extract and compose.}
We rank all candidate sections by their scores and select as many as the
far-memory budget allows. One- and two-chunk budgets allow $40$ and $80$
sections for T2V, and $48$ and $96$ for I2V, respectively. Their stored
token indices retrieve the verbatim KV entries at every backbone layer.

Softmax converts all candidate scores into weights, giving higher-scoring
sections larger weights. We normalize the selected weights to average $1$
and use them to scale each section's values. This differentiable weighting
lets the denoising loss train the descriptor encoder through the frozen DiT.

We preserve each selected token's spatial coordinates and adjust its
temporal RoPE coordinates to place it before the sliding window, fitting
the backbone's context layout. For $N$ selected sections, let $K_i,V_i$
denote section $i$'s KV after this positional adjustment, and $w_i$ its
value weight. At each backbone layer, the \method keys and values are
\begin{equation}
\begin{aligned}
K_{\mathrm{far}} &= \operatorname{Concat}(K_1,\ldots,K_N),\\
V_{\mathrm{far}} &= \operatorname{Concat}(w_1V_1,\ldots,w_NV_N).
\end{aligned}
\label{eq:supp_compose}
\end{equation}
Here $\operatorname{Concat}$ joins token rows. The frozen DiT reads the
resulting \method as far memory alongside the attention sink and sliding window.

\subsection{Self-Distillation}
\label{app:training}

We train the descriptor encoder through self-distillation to match the
same frozen generator's predictions under richer far memory.
Teacher and student use the same frozen DiT, noisy latent, denoising step, and sliding window; only
far memory differs.

\parnobf{Teacher and student.}
The KV context consists of an attention sink, a sliding window, and far
memory. We specify these components for each task below.

\emph{T2V.} Both teacher and student use text tokens as the attention sink
and the two most recent video chunks as the sliding window. The teacher
keeps the first five video chunks, which cover the prompt's initial reveal
of the target content. Their KV outside the sliding window forms its far
memory. The student's far memory is a \method composed
from historical sections outside the sliding window, under a two-chunk far-memory budget.

\emph{I2V.} Both teacher and student use the first video chunk as the
attention sink and the two most recent chunks as the sliding window.
The teacher's far memory contains up to three historical chunks nearest to
the current camera pose, selected from outside the attention sink and sliding window.
For the in-place training trajectory, nearness is measured by the angle
between chunks' mean viewing directions. The student's far memory is a
\method composed from historical sections outside the attention sink and sliding window,
under a two-chunk far-memory budget.

We minimize the mean squared error (MSE) between teacher and student
predictions, comparing noise-free latents for H3-AR and denoising velocities
for LingBot-World-Infinity. We scale both losses by $100$.
For T2V, we first generate a $16$-chunk rollout, then uniformly sample
one chunk from indices $5$--$15$ and one of its four denoising steps for
supervision. For I2V, we generate $20$ chunks sequentially. For each chunk
at indices $4$--$19$, we uniformly sample one of four denoising steps and
update the descriptor encoder at that step before continuing generation.
Each comparison uses the same noisy latent for teacher and student.
Chunk indices are zero-based.

Teacher predictions are fixed targets. Gradients pass through the student's
frozen DiT and the section value weights to the descriptor encoder;
they do not pass through the discrete section selection. The stored KV and
backbone parameters remain unchanged. The mean used to normalize value weights is held
constant during backpropagation, preserving gradients through the
full-candidate softmax to unselected scores as well.
To explore alternatives during training, random candidates replace $15\%$
of selected sections for T2V and a fraction drawn uniformly from
$[0,0.15]$ for I2V. Evaluation disables this exploration.

\parnobf{Training data.}
T2V uses $2000$ four-segment prompt sequences based on Ring Forcing's
three-stage appear--disappear--reappear design~\citep{xue2026ringforcingpreciselongterm},
with $16$ chunks per rollout. I2V starts from $2664$ Sekai
clips~\citep{li2025sekaivideodatasetworld}, captioned with
Qwen2.5-VL-7B~\citep{bai2025qwen25vltechnicalreport}. Scene-disjoint splits
yield $2388$ training clips and $44/46$ validation/test clips after capping
each held-out scene at three clips. Sekai supplies conditioning frames; the
frozen backbone generates continuations, so no ground-truth target videos
are required. Training uses a $20$-chunk in-place turn of roughly
$120^\circ$ followed by its reversal. The DL3DV evaluation scenes and
$90^\circ$, $180^\circ$, $360^\circ$, and translation trajectories are
separate from this training setup (Section~\ref{app:construction}).

\parnobf{Optimization.}
Table~\ref{tab:training} reports architecture, optimization, and checkpoint
settings. Each backbone has its own trained router, reused across memory
budgets. Scheduled steps specify the training horizon; evaluation uses
the earlier checkpoint listed in the table.

\begin{table}[!htb]
\centering
\caption{\textbf{Self-distillation configuration.} Only the router is
optimized. Scheduled steps describe the configured training schedule;
reported checkpoints identify the weights used for evaluation.}
\label{tab:training}
\small
\begin{tabular}{@{}lll@{}}
\toprule
 & T2V (H3-AR) & I2V (LingBot-World-Infinity) \\
\midrule
Hardware & $32$ NVIDIA H200 & $16$ NVIDIA H200 \\
Parallelism & FSDP, sequence parallel $4$ & DDP \\
Sequences per step & $8$ & $16$ \\
Trainable parameters & $13.65$\,M & $11.55$\,M \\
Optimizer & AdamW & AdamW \\
Learning rate & $10^{-4}$ & $2\times10^{-4}$ \\
Adam betas & $(0.9,0.95)$ & $(0.9,0.999)$ \\
Weight decay & $0.01$ & $0$ \\
Learning-rate schedule & Constant & $200$-step warmup, then cosine \\
Gradient clipping & $1.0$ & $1.0$ \\
Scheduled steps & $4000$ & $4768$ (two epochs) \\
Reported checkpoint & Step $416$ & Step $500$ \\
\midrule
Training data & $2000$ prompt sequences & $2388$ Sekai clips \\
Training rollout & $16$ chunks & $20$ chunks \\
Student far memory & $2$ chunks & $2$ chunks \\
Sliding window & $2$ video chunks & $2$ video chunks \\
Attention sink & Text tokens & First video chunk \\
Teacher historical chunks & First $5$ video chunks & $3$ chunks with closest poses \\
Descriptor width & $1024$ & $1024$ \\
Initial score multiplier, $\beta$ & $8$ (fixed) & $8$ (learned) \\
\bottomrule
\end{tabular}
\end{table}

%% file: sec/appendix/B_benchmarks.tex
\section{Benchmark Details}
\label{app:benchmarks}

This section describes the construction and evaluation of \bench{} and illustrates its visual diversity.
T2V test scenarios are disjoint from router training. I2V uses initial
frames from DL3DV rather than the Sekai frames used for training, together
with camera trajectories not seen during training.

\subsection{Dataset Construction}
\label{app:construction}
\label{app:dataset_construction}

\parnobf{T2V prompts and screening.}
\label{app:t2v_screen}
We adapt Ring Forcing's three-stage appear--disappear--reappear
design~\citep{xue2026ringforcingpreciselongterm} into four prompt segments with a Claude agent:
the object appears, disappears, remains out of sight, and reappears.
The candidate pool contains 466 scenes covering cases, drawers, doors,
lids, covers, tins, hinges, latches, screw tops, and sliding covers.
The nominal prompt transitions occur at 3.6, 6.2, and 11.2 seconds.
The third segment therefore requests five seconds with the object out of
sight, exceeding the largest sliding-window baseline's approximately
three seconds of recent context. Each generated clip contains 379 frames
at 24\,fps; the prompt schedule specifies the intended action timing,
not the exact frame at which a generated action occurs.

\begin{figure}[tb]
\centering
\includegraphics[width=0.96\textwidth]{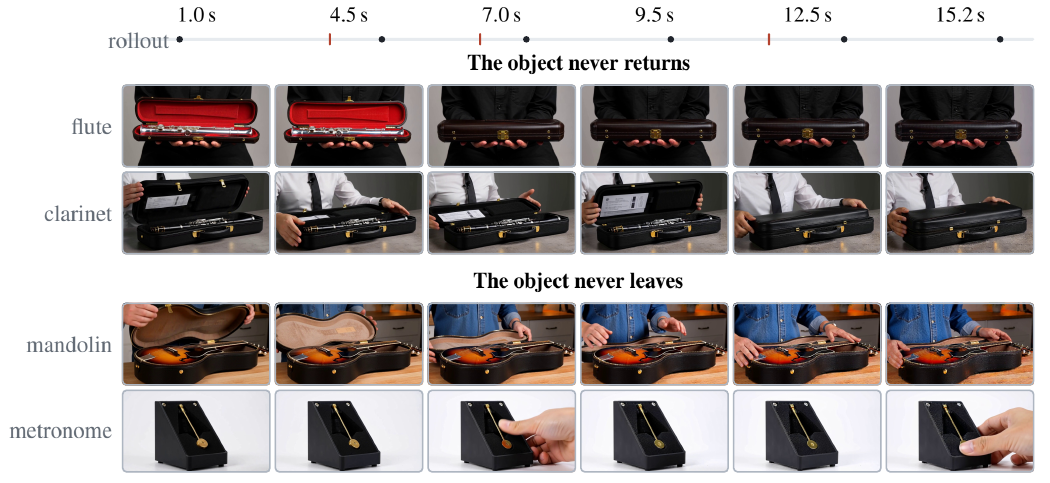}
\caption{\textbf{T2V prompt-compliance failures.} Each row shows a discarded
scene under sliding-window inference. The upper band marks the prompt
transitions and sampled instants. Top: the container closes but never
reopens. Bottom: the container never closes.}
\label{fig:follow}
\end{figure}

For each prompt, we generate one H3-AR rollout and screen it for compliance
with the four segments. We retain 100 scenes, excluding 366. A rollout in which the object never
disappears or never returns has no valid departure--revisit pair and does
not test the memory behavior of interest. Figure~\ref{fig:follow} shows
these two failure modes. The retained scene list and noise seeds are
fixed across the methods and budgets being compared.

\parnobf{I2V initial frames and prompts.}
\label{app:vlm}
We take initial frames from DL3DV~\citep{ling2023dl3dv10klargescalescenedataset},
center-crop them to the backbone's aspect ratio, and resize them to
$832\times480$. Qwen2.5-VL-7B~\citep{bai2025qwen25vltechnicalreport}
is used for scene classification, view screening, and captioning.
The classification pass separates indoor and outdoor scenes. The screening
pass checks eye-level views in both classes and overhead cover in outdoor
scenes only. A covered outdoor scene is eligible for rotation alone;
it is not used for translation. Ambiguous binary replies, containing both
or neither expected answer, are not assigned a default class.
The final set contains 50 indoor and 100 outdoor scenes; the selected
outdoor scenes support both trajectory types in
Table~\ref{tab:benchmark_settings}. The captioning
pass produces a scene prompt describing visible content without prescribing camera motion,
so the camera trajectory is supplied through the backbone's motion
conditioning rather than through the text prompt. Table~\ref{tab:vlm_prompts}
lists the scene-classification, eye-level screening, and captioning prompts.

\begin{table}[!htbp]
\centering
\small
\setlength{\tabcolsep}{6pt}
\renewcommand{\arraystretch}{1.16}
\caption{\textbf{Prompts for I2V scene preparation.} These prompts cover
scene classification, eye-level screening, and captioning for both indoor
and outdoor frames.}
\label{tab:vlm_prompts}
\begin{tabular}{@{}p{0.18\textwidth}p{0.77\textwidth}@{}}
\toprule
Purpose & Prompt \\
\midrule
Scene class & Is this an indoor scene or an outdoor scene? Answer with one
word: indoor or outdoor. \\
Eye-level view & Is this photo taken from roughly a standing person's eye level,
looking horizontally -- not from high above the scene, not tilted down at the
ground, not tilted up at the sky? Answer with one word: yes or no. \\
Scene prompt & Describe this scene in two or three sentences, as a caption for
the image. Name the place, the main structures and surfaces, the materials,
the lighting and the weather. Do NOT describe any camera motion, and do not
say 'the camera' or 'the video' -- describe only what is visible. \\
\bottomrule
\end{tabular}
\end{table}

\parnobf{I2V camera trajectories.}
All 150 scenes have in-place rotation trajectories, with a left or right
direction drawn once per scene and shared across settings. For
$90^\circ$ and $180^\circ$, yaw increases linearly to the specified angle
at the midpoint and then reverses to its initial value. The $360^\circ$
trajectory instead completes one continuous full turn.

For the 100 outdoor scenes, Depth Anything
3~\citep{lin2025depth3recoveringvisual} estimates depth from the initial
frame. We use conservative free-depth estimates around the forward
direction to define a scene-specific straight path into the scene and
back. The translation distance is bounded by the estimated free space. Rotation is superimposed on this out-and-back
translation, using the same three angular settings. Each input camera trajectory
ends at its initial position and orientation.

\FloatBarrier
\subsection{Visualization of First Frames}
\label{app:first_frames}

\begin{figure}[!htb]
\centering
\includegraphics[width=0.9\textwidth]{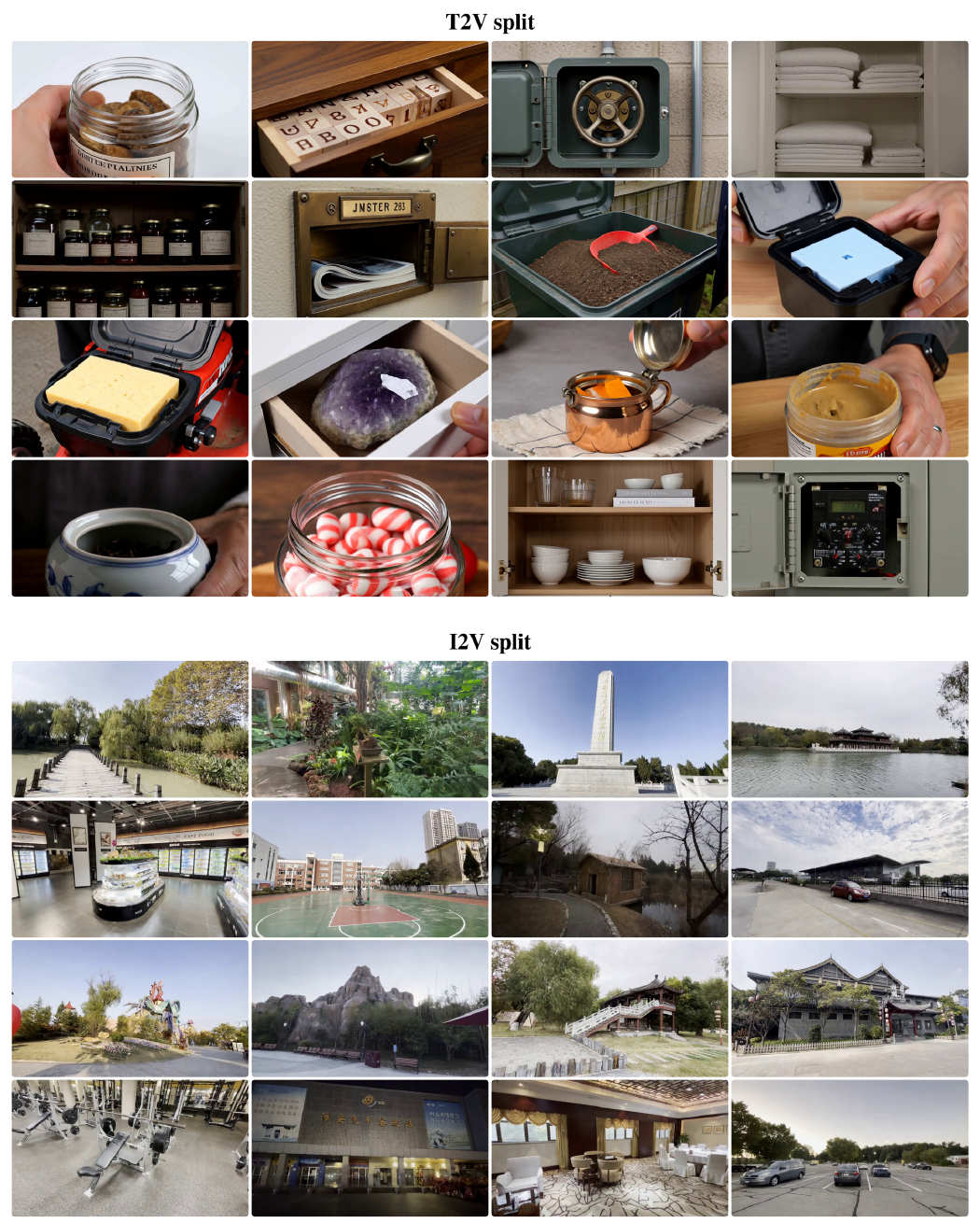}
\caption{\textbf{Example frames from the two splits.} Top: model-generated
frames from the T2V split when the prompt first reveals the object. Bottom:
conditioning frames from the I2V split, taken from the first frames of DL3DV
videos. Sixteen scenes are sampled at random from each split.}
\label{fig:bench}
\end{figure}

Figure~\ref{fig:bench} shows objects when first revealed in T2V rollouts
and the conditioning frames for I2V, illustrating the visual diversity
of the two splits.

\subsection{Detailed Evaluation Protocol}
\label{app:evaluation}
\label{app:marking}

\parnobf{Generation and matched budgets.}
To isolate memory quality, all methods use the same prompts and noise seed
for each scene, with a shared prompt schedule for T2V and the same
conditioning frame and input camera trajectory for I2V.
T2V generates $379$ frames at $24$\,fps ($23$ chunks); I2V generates
$253$ frames at $16$\,fps ($16$ chunks). Both use four denoising steps per
chunk. Retrieval methods retain two chunks in the sliding window plus one or two chunks
of far memory. Base instead retains three or four recent chunks, matching
the active visual KV budget. All methods use the same backbone-specific
attention sink. Scores are computed on the original rollouts, before the
compression used for the \href{https://mosaichunk.github.io/viewer/video.html}{Video~Viewer}.

\parnobf{MoC adaptation.}
We adapt MoC~\citep{cai2025mixturecontextslongvideo} to our frozen backbones,
using mean-pooled keys to describe and retrieve historical chunks.
We use the latest chunk's
mean-pooled section keys as retrieval queries. Each historical chunk is
described by its mean-pooled keys and scored by its maximum dot product
with these queries. The highest-scoring one or two chunks form far memory,
with the same selection shared across queries, attention heads, and layers.
Retrieved values have unit weights, and positional handling is the same
as for \method.

\begin{table}[!htbp]
\centering
\small
\setlength{\tabcolsep}{5pt}
\renewcommand{\arraystretch}{1.12}
\caption{\textbf{\bench{} evaluation settings.} The same inputs are used for
all methods. Outdoor I2V scenes support both trajectory types.}
\label{tab:benchmark_settings}
\begin{tabular*}{\textwidth}{@{\extracolsep{\fill}}lll@{}}
\toprule
 & T2V & I2V \\
\midrule
Scenes & 100 & 50 indoor + 100 outdoor \\
Backbone & H3-AR & LingBot-World-Infinity \\
Resolution & $1376\times768$ & $832\times480$ \\
Frames / frame rate & 379 / 24\,fps & 253 / 16\,fps \\
Revisit control & Four-segment prompt & Camera trajectory \\
Rotation & --- & $90^\circ$, $180^\circ$, $360^\circ$; 150 scenes \\
Rotation + translation & --- & Same angles; 100 outdoor scenes \\
Departure frame & Manually marked & Conditioning frame \\
Revisit frame & Manually marked per rollout & Selected from Pi3X poses \\
\bottomrule
\end{tabular*}
\end{table}

\begin{figure}[tb]
\centering
\includegraphics[width=0.94\textwidth]{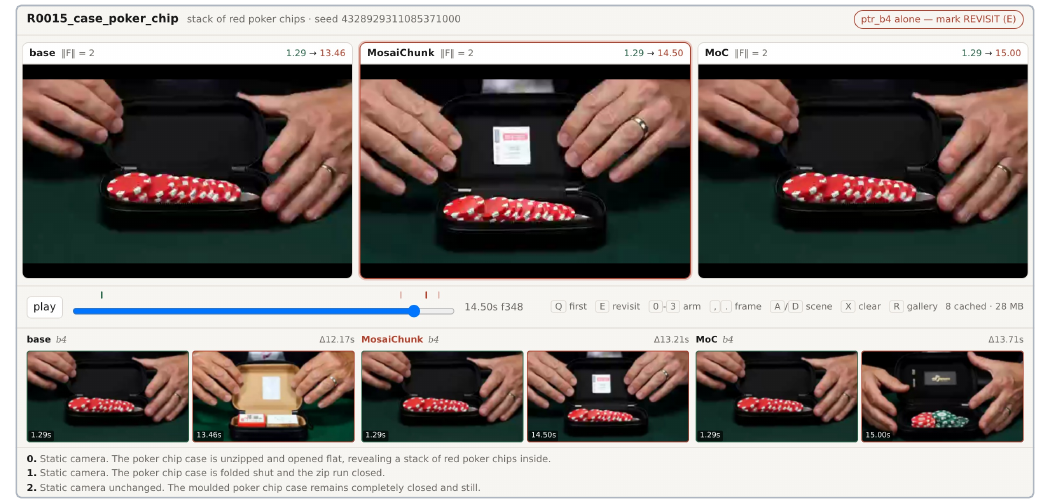}
\caption{\textbf{T2V frame annotation.} During revisit annotation, the
timeline controls only the selected rollout (red outline); the other
videos remain paused. Q marks the shared departure frame, and E marks
the selected rollout's revisit frame. Saved departure--revisit pairs
appear below the timeline.}
\label{fig:viewer}
\end{figure}

\parnobf{T2V frame pairs.}
For each retained scene, we manually mark a departure frame in which the
object is fully visible before the container closes. Its frame index is
shared across methods. We then mark each rollout's revisit frame
separately, when the container has reopened and its contents are visible.
This accommodates differences in generated action timing. The annotation interface in Figure~\ref{fig:viewer} uses linked playback
to mark the common departure frame. We then select and scrub each rollout
individually to mark its revisit frame; the saved frame pairs appear below.

\parnobf{I2V frame pairs.}
The departure frame is the conditioning frame. We reconstruct each
generated rollout with Pi3X~\citep{wang2026pi3permutationequivariantvisualgeometry}
in a single pass, using every decoded frame. Reconstruction inputs are
resized to approximately 255,000 pixels, with dimensions rounded to
multiples of 14. The reconstructed poses determine the revisit frame independently
for each method.

For rotation, the turnaround is the frame with the largest angular
deviation of its viewing direction from that of the conditioning frame. For
translation, it is the frame farthest from the initial position.
Among frames from the turnaround onward, we select the revisit frame with
the smallest unsigned angle between its reconstructed viewing direction
and that of the conditioning frame. This rule also handles a full $360^\circ$ turn without an
angle-wrap ambiguity.
CLIP and LPIPS are evaluated on the selected original images.

\parnobf{Metrics and aggregation.}
Table~\ref{tab:revisit_metrics} specifies the metrics used for the
T2V and I2V comparisons. CLIP and LPIPS compare the departure--revisit pair;
TempSSIM and Drift summarize the full rollout. For the latter two, we
first average within each video, then report the median over scenes,
as for the pairwise metrics. We use the same definitions for every
method and memory budget.

\begin{table}[!htbp]
\centering
\small
\setlength{\tabcolsep}{6pt}
\renewcommand{\arraystretch}{1.16}
\caption{\textbf{Metrics on \bench.} CLIP and LPIPS assess revisit
consistency; TempSSIM and Drift provide complementary measures of rollout
quality. Arrows indicate the preferred direction.}
\label{tab:revisit_metrics}
\begin{tabular}{@{}p{0.18\textwidth}p{0.77\textwidth}@{}}
\toprule
Metric & Definition \\
\midrule
CLIP $\uparrow$ & Cosine similarity of L2-normalized image embeddings
from CLIP ViT-H/14, using the LAION-2B checkpoint and its standard image
processor. \\
LPIPS $\downarrow$ & LPIPS with AlexNet (version 0.1), evaluated at each
backbone's native image resolution after scaling pixels to $[-1,1]$. \\
TempSSIM $\uparrow$ & Mean SSIM over all consecutive decoded frame pairs
in grayscale, with an $11\times11$ Gaussian window and $\sigma=1.5$. \\
Drift $\downarrow$ & Mean cosine distance between adjacent chunks.
Each chunk is represented by the normalized average of CLIP embeddings
from four evenly spaced frames within that chunk. \\
\bottomrule
\end{tabular}
\end{table}

For Drift, chunk boundaries follow the backbone's decoded output:
H3-AR has an initial five-frame chunk followed by 17-frame chunks;
LingBot-World-Infinity has an initial 13-frame chunk followed by
16-frame chunks. This avoids treating an arbitrary fixed-length frame
partition as the model's chunk structure.
\FloatBarrier

%% file: sec/appendix/C_results.tex
\FloatBarrier
\section{Additional Results}
\label{app:results}

\subsection{Extra Qualitative Results}
\label{app:qual}

\begin{figure}[!htb]
\centering
\includegraphics[width=\textwidth]{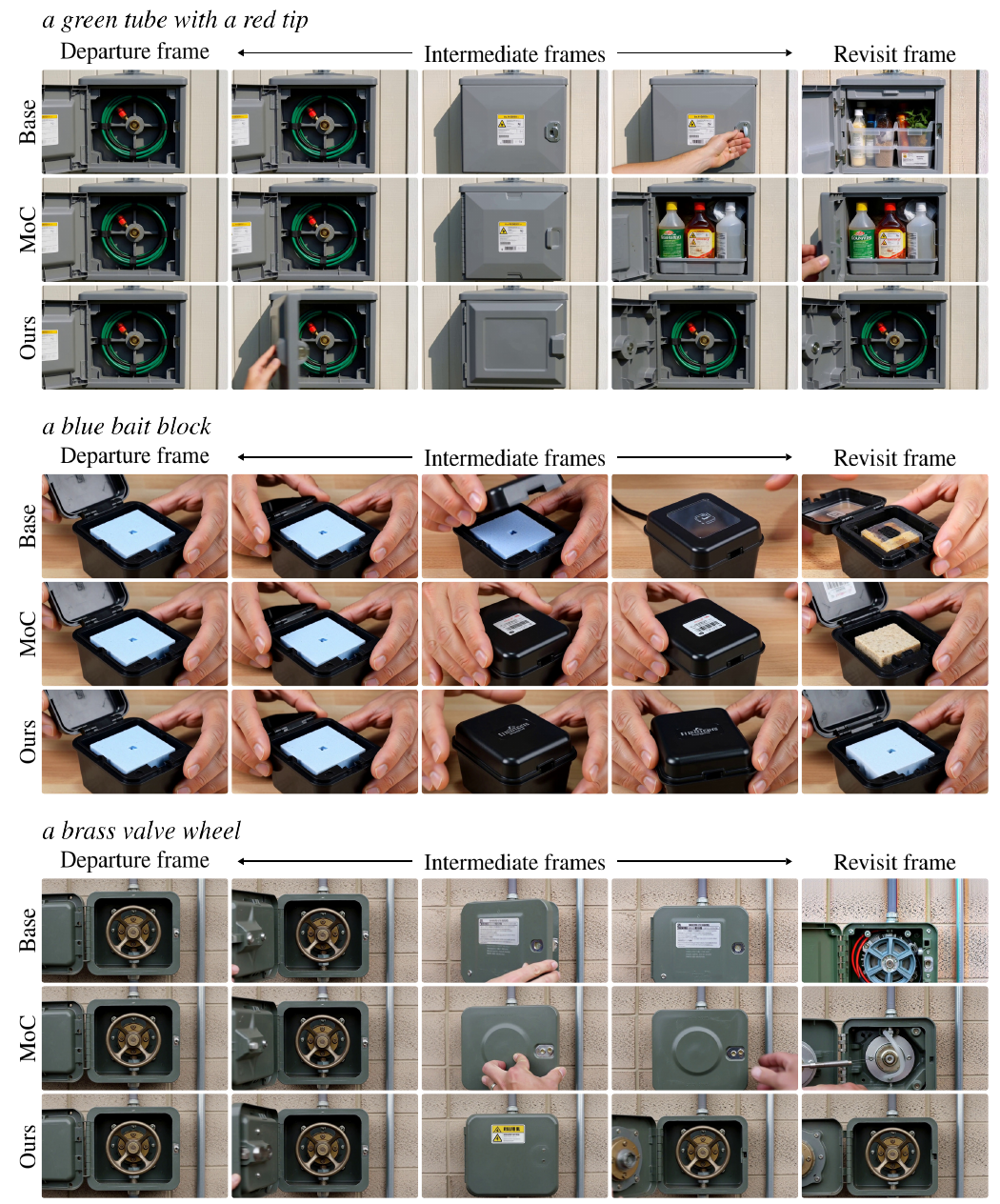}
\caption{\textbf{Additional T2V revisits at a two-chunk far-memory budget.}
The first column shows the departure frame before the target content
leaves view; the last shows each method's marked revisit frame after it
returns. Intermediate columns sample the rollout. Ours denotes \method.}
\label{fig:qual_supp}
\end{figure}

\begin{figure}[!htb]
\centering
\includegraphics[width=\textwidth]{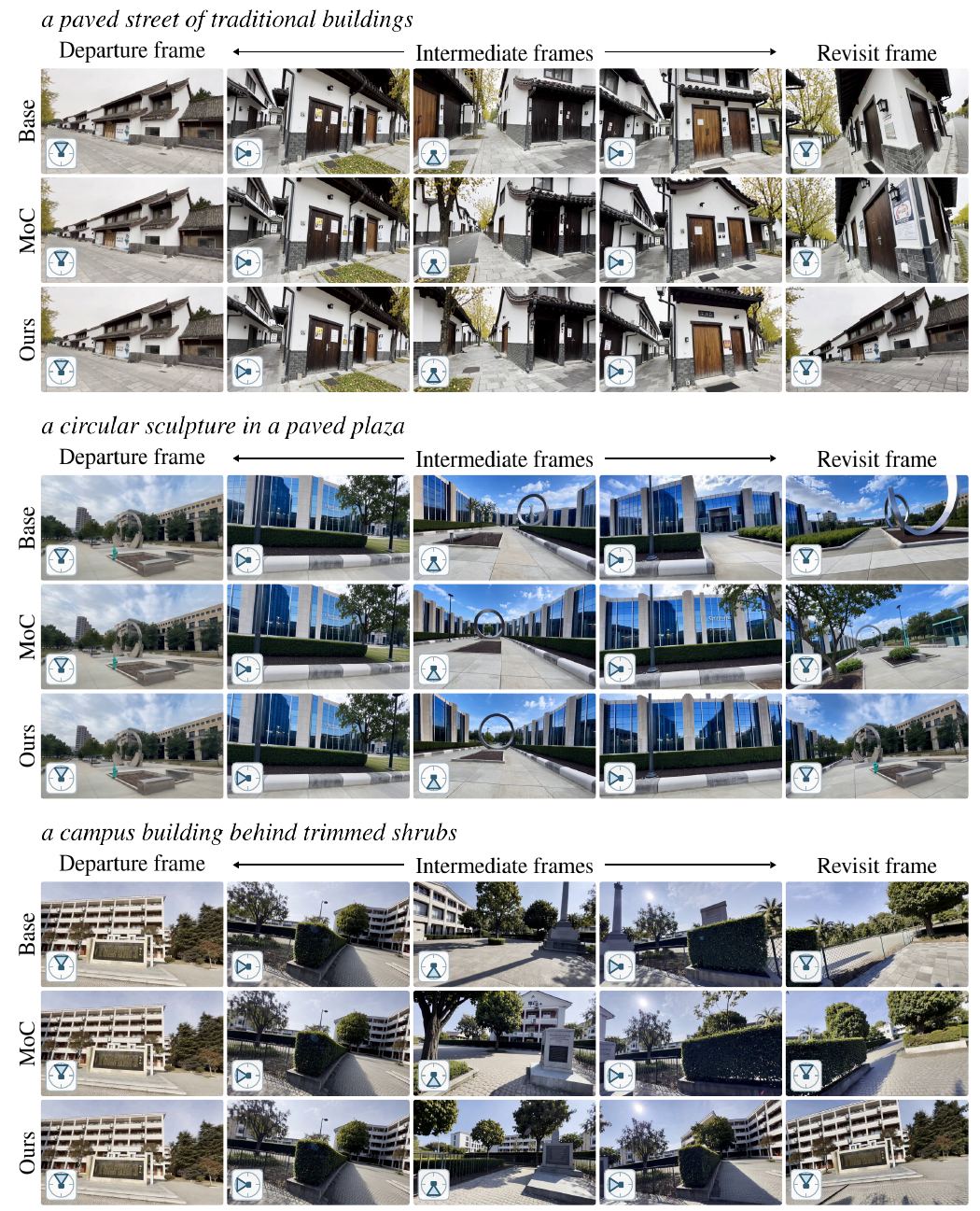}
\caption{\textbf{Additional I2V revisits at a one-chunk far-memory budget.}
All methods start from the same conditioning frame and follow the same
$180^\circ$ trajectory. The first column shows the conditioning frame,
the intermediate columns sample the camera turn, and the last shows
each method's selected revisit frame. Camera insets schematically show
the commanded rotation. Ours denotes \method.}
\label{fig:qual_i2v_supp}
\end{figure}

Figures~\ref{fig:qual_supp} and~\ref{fig:qual_i2v_supp} show three
scenes per setting, using the frame-pair protocol in
Section~\ref{app:marking}. The first and last columns show departure and revisit
frames, with intermediate frames illustrating the intervening motion.

Within each scene, methods
share the prompt and noise seed; I2V methods also share the conditioning
frame and input camera trajectory. Rows therefore differ only in the memory
supplied to the backbone. Rows are Base, MoC, and Ours (\method), from top
to bottom. The T2V examples use two chunks of far memory; the I2V examples
use one, as stated in the captions.
The \href{https://mosaichunk.github.io/viewer/video.html}{Video~Viewer} provides synchronized full rollouts together with
the departure--revisit frame pairs and their CLIP scores.

\FloatBarrier

\subsection{Extra Quantitative Results}
\label{app:quantitative}
\label{app:public}

\parnobf{Public benchmarks.}
We evaluate I2V generation on WBench~\citep{ying2026wbench} and
WorldMark~\citep{xu2026worldmark}, two general-purpose benchmarks for
interactive video world models, and report every metric in their
consistency and video-quality dimensions. We compare Base and \method at one-
and two-chunk far-memory budgets. Base matches the total active KV cache
budget by retaining additional recent chunks instead of far memory.
Scores are reported on a $0$--$100$ scale, with higher values indicating
better performance. Both benchmarks use
the frozen LingBot-World-Infinity backbone at $832\times480$ and $16$\,fps,
with four denoising steps per chunk and identical noise seeds across
compared methods.

Table~\ref{tab:public_benchmarks} summarizes selected metrics; full results
and metric descriptions follow.

\ifdefined\standalonesupplement
\input{tables/public_benchmarks}
\fi

The relatively small camera rotations in these evaluations can preserve
substantial overlap with earlier views, as in our $90^\circ$ setting,
reducing the need to recall content beyond the sliding window.
The sliding-window baseline already achieves strong consistency scores
in these evaluations.
Nevertheless, \method improves every consistency metric reported in
Table~\ref{tab:public_benchmarks} at both budgets, while also
improving the video-quality scores.

\parnobf{WBench.}
We use the navigation split with $158$ cases. Spatial and Gated Spatial
are defined on its $60$ round-trip cases, and Subject is evaluated on
$103$ cases; the remaining metrics use all $158$ cases.
The released adapter converts navigation commands into camera
poses and combines segment prompts while preserving perspective information
and command durations. Table~\ref{tab:wbench_supp} reports the full results. The metric
descriptions below follow the released evaluation implementation.

For \emph{consistency}, \emph{Background} compares CLIP features across
time. \emph{Spatial} compares the initial and returning views using
DreamSim, with the return frame selected from estimated camera
orientations. \emph{Gated Spatial} downweights this similarity when
intermediate frames show little departure from the initial view.
\emph{Segment} reports the fraction of videos without detected shot
boundaries. \emph{Perspective} measures the stability and presence of a
tracked subject's image-plane centroid. \emph{Subject} compares masked
subject features across adjacent frames and against the first frame.
\emph{Geometric} measures agreement between estimated depths after
reprojection, and \emph{Photometric} measures RGB agreement under the same
geometric warping.

For \emph{video quality}, \emph{Aesthetic} uses a learned aesthetic
predictor and \emph{Imaging} uses MUSIQ to assess technical image quality.
\emph{Flickering} penalizes pixel changes between successive frames,
\emph{Dynamic} detects motion from optical flow, and \emph{Smoothness}
evaluates frame-interpolation error. \emph{HPSv3-Norm} reports
percentile-normalized human-preference scores.

\input{tables/wbench_supp}

Table~\ref{tab:wbench_supp} shows higher Background, Spatial, Gated Spatial,
Perspective, Subject, and Geometric scores for \method at both budgets.
The improvements are not uniform across all metrics: Segment is lower at
two chunks, and Photometric is slightly lower at that budget.

\parnobf{WorldMark.}
We evaluate the Real and Stylized first-person splits, each containing
$125$ videos per method and budget, and macro-average the scores across
the two splits. Table~\ref{tab:worldmark_supp} gives the complete results
for the World Memory and Visual Quality dimensions. We use the supplied
conditioning frames, prompts, and image-specific intrinsics, adjusted after
center cropping. Each action segment lasts $20$ seconds and commands
$120^\circ$ of rotation or two world units of translation. Completing the
final generation chunk produces $333$, $653$, or $973$ frames for one-,
two-, or three-segment trajectories, respectively.

\emph{Local Memory} penalizes detected cuts and abrupt changes in DINOv2
features between frames sampled at equal increments of accumulated
motion. \emph{Global Memory} measures geometric consistency across the
trajectory using VGGT-Omega reconstruction. \emph{Revisit Memory} aligns
outward and return frames by cumulative optical-flow arc length and
compares DINOv2 features of the matched pairs, using clips that satisfy
the benchmark's return-motion checks. \emph{Aesthetic} and
\emph{Perceptual} use the aesthetics and visual-quality tasks of Q-Align,
respectively.

\input{tables/worldmark_supp}
Table~\ref{tab:worldmark_supp} shows improved Global and Revisit Memory at
both budgets, alongside higher Aesthetic and Perceptual scores. Local
Memory is slightly lower, distinguishing improved recall from local
temporal stability.
\FloatBarrier

\parnobf{Additional camera trajectories.}
\label{app:angles}
Table~\ref{tab:i2v_supp} reports $90^\circ$ and $360^\circ$ turns, with and
without translation, using the trajectories in Section~\ref{app:construction}
and frame-pair selection in Section~\ref{app:marking}.
At $90^\circ$, the starting view remains substantially
visible and the methods achieve similar scores. At $360^\circ$, revisit
consistency is lower for all methods and their relative ordering varies
across settings. Thus, the benefits observed for the $180^\circ$ revisit
task do not extend uniformly to every camera trajectory: limited turns
preserve overlapping visual context, while full rotations remain
challenging.

\input{tables/i2v_supp}
\FloatBarrier

\subsection{Extra Ablations}
\label{app:ablation}

\parnobf{Per-query selection details.}
For the per-query selection ablation, we retrain the descriptor encoder to
select historical sections independently for each query section in the
latest chunk. Selecting one or two sections per query gives the one- or
two-chunk far-memory budget, respectively.

\parnobf{Camera pose as a retrieval signal.}
We examine whether camera poses improve descriptor-based retrieval and
whether section composition remains useful when this guidance is
available. \method scores historical sections using learned
descriptor similarity. \methodpose first uses camera-pose similarity to
shortlist historical chunks, then applies the same trained router to
select sections within them. For a whole-chunk comparison, we implement
WorldKV's pose-based retrieval rule~\citep{yi2026worldkv} on the same frozen
backbone. Table~\ref{tab:i2v_pose} evaluates $180^\circ$ trajectories on
$150$ rotation scenes and $100$ scenes with rotation and translation, at one- and
two-chunk far-memory budgets.

\input{tables/i2v_pose}

\begin{figure}[tb]
\centering
\includegraphics[width=0.85\textwidth]{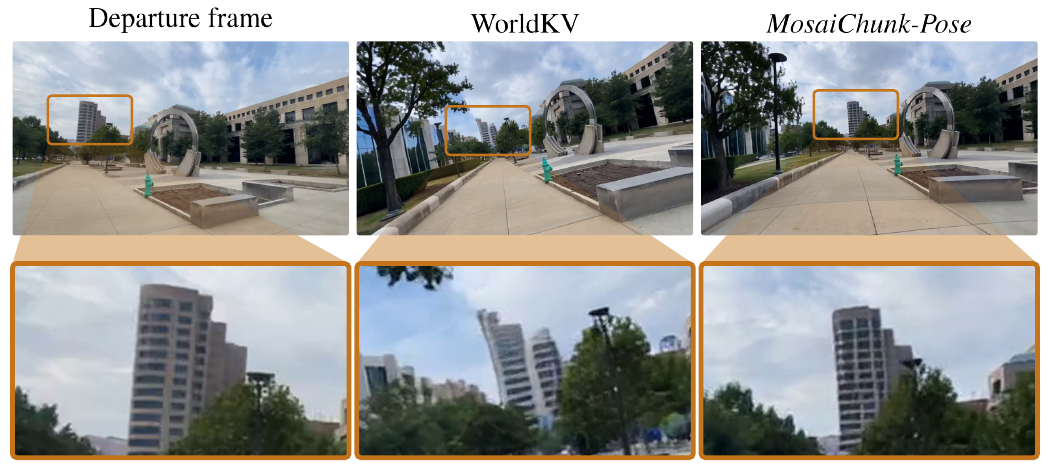}
\caption{\textbf{Pose-guided retrieval at a one-chunk far-memory budget.}
The left panel is the conditioning frame. Compared with WorldKV,
\methodpose better preserves the background building's staircase.}
\label{fig:pose_i2v}
\end{figure}

Pose guidance improves both CLIP and LPIPS over \method in
every setting. Relative to WorldKV, \methodpose achieves similar CLIP,
with slightly higher scores at both two-chunk budgets. WorldKV is stronger
on rotation at one chunk and has lower LPIPS under rotation with
translation. Figure~\ref{fig:pose_i2v} illustrates how section composition
can preserve the staircase of a background building, even though it
occupies only a small part of the frame.

\begin{figure}[!htb]
\centering
\includegraphics[width=\textwidth]{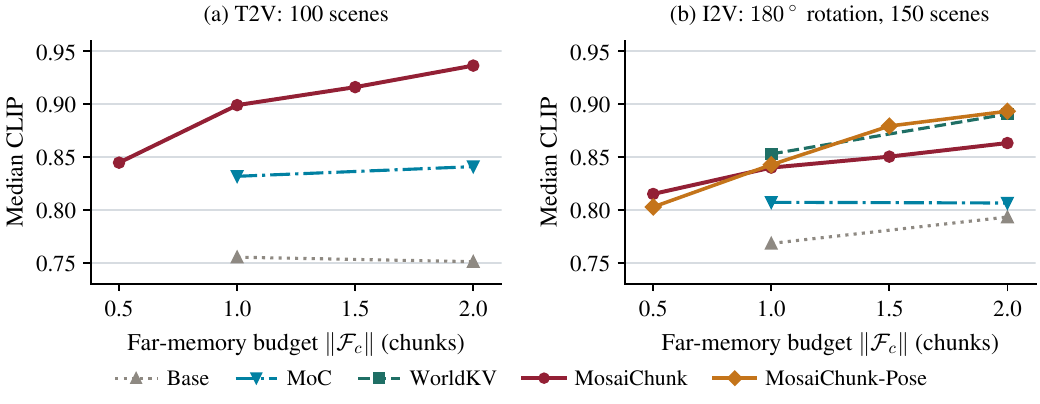}
\caption{\textbf{Revisit consistency across far-memory budgets.}
Median CLIP on $100$ T2V scenes and $150$ I2V $180^\circ$ rotation scenes.
\method and \methodpose use half-chunk increments; other methods use
one- and two-chunk budgets. Base matches the total active KV cache
budget by retaining additional recent chunks.}
\label{fig:budget}
\end{figure}

\parnobf{Far-memory budget.}
We compare allocating additional KV entries to retrieved far memory
with retaining additional recent chunks. Retrieval methods keep their
sliding window fixed, whereas Base matches the total active KV cache budget
by retaining additional recent chunks. On $100$ T2V and $150$ I2V rotation scenes,
we evaluate \method and, for I2V, \methodpose at far-memory budgets from $0.5$ to $2$ chunks in
half-chunk increments. The other methods are evaluated at one and two
chunks (Figure~\ref{fig:budget}).

Both \method and \methodpose improve as the budget grows. The advantage
also holds with less far memory: on T2V, \method with one chunk exceeds
MoC with two ($0.899$ vs.\ $0.841$ CLIP); on I2V, half a chunk already
does so ($0.815$ vs.\ $0.807$). These comparisons demonstrate both
effective retrieval at small budgets and improved revisit consistency
when more memory is available.

\FloatBarrier

\subsection{KV Transplantation}
\label{app:kv_reuse}

\begin{figure}[!htb]
\centering
\includegraphics[width=0.68\textwidth]{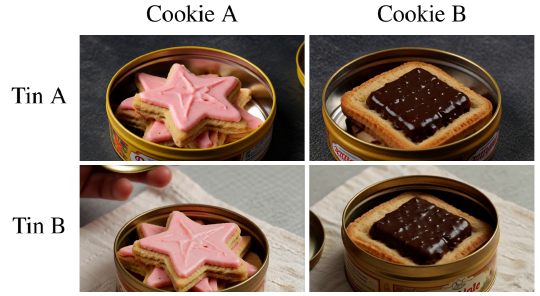}
\caption{\textbf{Transplanting content between rollouts.}
Rows specify the tin and columns the cookie. Diagonal images are source
frames; off-diagonal images show the swapped results.}
\label{fig:swap}
\end{figure}

We test whether selected historical KV can transfer content between
rollouts without updating the video generator.
We generate two cookie-tin rollouts with different cookies, manually mark
each cookie in an earlier frame, and select the cached KV at the
corresponding latent positions. When the tins close and reopen, we
exchange these entries as far memory while keeping the generator frozen.
Figure~\ref{fig:swap} shows that cookie A appears in tin B and cookie B
appears in tin A, while each rollout retains its original tin. This
intervention provides further evidence that historical KV can serve as
content-specific memory even when it comes from another rollout.

\FloatBarrier

%% file: tables/wbench_supp.tex
\begin{table}[!ht]
\centering
\small
\setlength{\tabcolsep}{4pt}
\caption{\textbf{Full WBench results.} Consistency and Video Quality
metrics on the navigation split: Spatial and Gated Spatial use $60$
round-trip cases, Subject uses $103$ cases, and the remaining metrics use
all $158$ cases. Budgets specify \method's far memory in chunk equivalents;
Base matches the total active KV cache budget by retaining additional recent chunks.
Higher scores are better. The higher consistency score at each budget is
bold, except ties; video-quality scores are not bolded.}
\label{tab:wbench_supp}
\begin{tabular}{l@{\hskip 9pt}cc@{\hskip 9pt}cc}
\toprule
 & \multicolumn{2}{c@{\hskip 9pt}}{1 chunk} & \multicolumn{2}{c}{2 chunks} \\
\cmidrule(l{2pt}r{8pt}){2-3}\cmidrule(l{2pt}r{2pt}){4-5}
Metric & Base & \method & Base & \method \\
\midrule
\multicolumn{5}{l}{\textit{Consistency}} \\
\quad Background     & 91.33 & \textbf{91.57} & 91.44 & \textbf{91.52} \\
\quad Spatial        & 78.28 & \textbf{79.63} & 79.43 & \textbf{80.73} \\
\quad Gated Spatial  & 75.47 & \textbf{76.69} & 76.78 & \textbf{78.18} \\
\quad Segment        & 97.47 & 97.47 & \textbf{97.47} & 94.94 \\
\quad Perspective    & 80.21 & \textbf{82.11} & 80.59 & \textbf{82.14} \\
\quad Subject        & 88.13 & \textbf{88.36} & 88.10 & \textbf{88.38} \\
\quad Geometric      & 85.98 & \textbf{86.74} & 86.14 & \textbf{87.08} \\
\quad Photometric    & 79.62 & \textbf{79.67} & \textbf{79.85} & 79.83 \\
\multicolumn{5}{l}{\textit{Video Quality}} \\
\quad Aesthetic      & 61.75 & 61.86 & 61.76 & 61.92 \\
\quad Imaging        & 67.31 & 67.42 & 67.40 & 67.45 \\
\quad Flickering     & 91.32 & 91.30 & 91.37 & 91.28 \\
\quad Dynamic        & 94.94 & 95.57 & 96.20 & 96.20 \\
\quad Smoothness     & 96.40 & 96.44 & 96.45 & 96.42 \\
\quad HPSv3-Norm     & 70.31 & 70.42 & 70.47 & 70.70 \\
\bottomrule
\end{tabular}
\end{table}

%% file: tables/worldmark_supp.tex
\begin{table}[!ht]
\centering
\small
\setlength{\tabcolsep}{4pt}
\caption{\textbf{Full WorldMark results.} All World Memory and Visual
Quality metrics, macro-averaged over the Real and Stylized first-person
splits, with $125$ videos per method and budget in each split. Budgets
specify \method's far memory in chunk equivalents; Base matches the total
active KV cache budget by retaining additional recent chunks. Higher scores are better.
The higher consistency score at each budget is bold; video-quality scores
are not bolded.}
\label{tab:worldmark_supp}
\begin{tabular}{l@{\hskip 9pt}cc@{\hskip 9pt}cc}
\toprule
 & \multicolumn{2}{c@{\hskip 9pt}}{1 chunk} & \multicolumn{2}{c}{2 chunks} \\
\cmidrule(l{2pt}r{8pt}){2-3}\cmidrule(l{2pt}r{2pt}){4-5}
Metric & Base & \method & Base & \method \\
\midrule
\multicolumn{5}{l}{\textit{Consistency}} \\
\quad Local Memory     & \textbf{95.55} & 95.29 & \textbf{93.70} & 93.46 \\
\quad Global Memory    & 62.15 & \textbf{62.22} & 61.75 & \textbf{63.00} \\
\quad Revisit Memory   & 80.14 & \textbf{84.89} & 84.63 & \textbf{84.87} \\
\multicolumn{5}{l}{\textit{Video Quality}} \\
\quad Aesthetic        & 61.43 & 62.48 & 61.05 & 62.89 \\
\quad Perceptual       & 85.49 & 86.45 & 84.97 & 86.64 \\
\bottomrule
\end{tabular}
\end{table}

%% file: tables/i2v_supp.tex
\begin{table}[!htb]
\centering
\small
\setlength{\tabcolsep}{3pt}
\renewcommand{\arraystretch}{1.08}
\caption{\textbf{I2V results at $90^\circ$ and $360^\circ$.}
Median revisit CLIP and LPIPS on $150$ rotation scenes and $100$ scenes
with rotation and translation. Retrieval methods use one- or two-chunk
far-memory budgets with a fixed sliding window; Base matches the total
active KV cache budget by retaining additional recent chunks. WorldKV and \methodpose use camera
poses; MoC scores pooled keys, while \method uses learned section descriptors.}
\label{tab:i2v_supp}
\begin{tabular*}{\textwidth}{@{\extracolsep{\fill}}llccccc@{}}
\toprule
\multicolumn{7}{c}{\textbf{CLIP $\uparrow$}} \\
\addlinespace[2pt]
Trajectory & Budget & Base & MoC & \method & WorldKV & \methodpose \\
\midrule
\multirow{2}{*}{$90^\circ$} & 1 chunk & 0.910 & 0.910 & 0.913 & 0.915 & 0.912 \\
 & 2 chunks & 0.907 & 0.908 & 0.906 & 0.915 & 0.910 \\
\addlinespace[2pt]
\multirow{2}{*}{$90^\circ$ + translation} & 1 chunk & 0.882 & 0.884 & 0.889 & 0.900 & 0.896 \\
 & 2 chunks & 0.881 & 0.875 & 0.887 & 0.895 & 0.889 \\
\addlinespace[2pt]
\multirow{2}{*}{$360^\circ$} & 1 chunk & 0.716 & 0.722 & 0.717 & 0.728 & 0.704 \\
 & 2 chunks & 0.711 & 0.719 & 0.719 & 0.714 & 0.717 \\
\addlinespace[2pt]
\multirow{2}{*}{$360^\circ$ + translation} & 1 chunk & 0.728 & 0.739 & 0.730 & 0.758 & 0.727 \\
 & 2 chunks & 0.733 & 0.726 & 0.752 & 0.730 & 0.748 \\
\midrule
\multicolumn{7}{c}{\textbf{LPIPS $\downarrow$}} \\
\addlinespace[2pt]
Trajectory & Budget & Base & MoC & \method & WorldKV & \methodpose \\
\midrule
\multirow{2}{*}{$90^\circ$} & 1 chunk & 0.530 & 0.536 & 0.538 & 0.526 & 0.529 \\
 & 2 chunks & 0.554 & 0.551 & 0.551 & 0.529 & 0.542 \\
\addlinespace[2pt]
\multirow{2}{*}{$90^\circ$ + translation} & 1 chunk & 0.564 & 0.571 & 0.563 & 0.529 & 0.530 \\
 & 2 chunks & 0.558 & 0.585 & 0.573 & 0.544 & 0.553 \\
\addlinespace[2pt]
\multirow{2}{*}{$360^\circ$} & 1 chunk & 0.691 & 0.688 & 0.694 & 0.693 & 0.686 \\
 & 2 chunks & 0.698 & 0.687 & 0.691 & 0.690 & 0.693 \\
\addlinespace[2pt]
\multirow{2}{*}{$360^\circ$ + translation} & 1 chunk & 0.693 & 0.676 & 0.682 & 0.683 & 0.668 \\
 & 2 chunks & 0.681 & 0.673 & 0.677 & 0.673 & 0.675 \\
\bottomrule
\end{tabular*}
\end{table}

%% file: tables/i2v_pose.tex
\begin{table}[!htb]
\centering
\small
\setlength{\tabcolsep}{5pt}
\renewcommand{\arraystretch}{1.08}
\caption{\textbf{Effect of pose-guided retrieval.}
Revisit CLIP and LPIPS (medians) and rollout-quality metrics on $150$
rotation scenes and $100$ scenes with rotation and translation, all with $180^\circ$
turns. WorldKV and \methodpose use camera poses; \method uses learned
section descriptors without pose guidance. Far-memory budgets are measured
in original video-chunk equivalents, with the sliding window fixed. Better CLIP
and LPIPS among the two pose-guided methods are bold.}
\label{tab:i2v_pose}
\begin{tabular*}{\textwidth}{@{\extracolsep{\fill}}lcccc@{}}
\toprule
Method & CLIP $\uparrow$ & LPIPS $\downarrow$ & TempSSIM $\uparrow$ & Drift $\downarrow$ \\
\midrule
\multicolumn{5}{@{}l}{\textbf{Far memory: 1 chunk}} \\
\addlinespace[2pt]
\multicolumn{5}{@{}l}{\textit{Rotation}} \\
\method & 0.840 & 0.625 & 0.456 & 0.051 \\
WorldKV & \textbf{0.853} & \textbf{0.610} & 0.458 & 0.051 \\
\methodpose & 0.843 & 0.624 & 0.459 & 0.051 \\
\addlinespace[2pt]
\multicolumn{5}{@{}l}{\textit{Rotation + translation}} \\
\method & 0.848 & 0.606 & 0.519 & 0.044 \\
WorldKV & 0.880 & \textbf{0.543} & 0.518 & 0.043 \\
\methodpose & \textbf{0.881} & 0.573 & 0.520 & 0.045 \\
\midrule
\multicolumn{5}{@{}l}{\textbf{Far memory: 2 chunks}} \\
\addlinespace[2pt]
\multicolumn{5}{@{}l}{\textit{Rotation}} \\
\method & 0.863 & 0.609 & 0.458 & 0.051 \\
WorldKV & 0.891 & 0.578 & 0.461 & 0.050 \\
\methodpose & \textbf{0.893} & \textbf{0.577} & 0.461 & 0.050 \\
\addlinespace[2pt]
\multicolumn{5}{@{}l}{\textit{Rotation + translation}} \\
\method & 0.862 & 0.584 & 0.513 & 0.046 \\
WorldKV & 0.897 & \textbf{0.542} & 0.519 & 0.045 \\
\methodpose & \textbf{0.899} & 0.546 & 0.514 & 0.047 \\
\bottomrule
\end{tabular*}
\end{table}